%% file: _main.tex
\documentclass{article} % For LaTeX2e
\usepackage{iclr2027_conference,times}

\input{math_commands.tex}

\usepackage{hyperref}
\usepackage{url}
\usepackage{booktabs}
\usepackage{multirow}
\usepackage{graphicx}
\usepackage{wrapfig}
\usepackage{algorithm}
\usepackage{algpseudocode}
\usepackage{microtype}
\usepackage{xcolor}
\definecolor{rebuttalblue}{RGB}{0,70,160}
\newcommand{\revised}[1]{\textcolor{black}{#1}}
\newcommand{\revisioncolor}{\color{black}}
\usepackage{enumitem}
\usepackage{amsmath}
\usepackage{amssymb}
\usepackage{pifont}
\usepackage{tikz}
\usetikzlibrary{arrows.meta,calc}

\title{\ours: Adaptive Attention Switching for Efficient Reasoning}

\author{
\normalfont
\begin{tabular}[t]{@{}p{0.48\textwidth}@{}p{0.48\textwidth}@{}}
\textbf{Anna Kuzina} &
\textbf{Mart van Baalen} \\
Qualcomm AI Research\textsuperscript{$\ddagger$} &
Qualcomm AI Research \\
{\scriptsize\texttt{akuzina@qti.qualcomm.com}} &
{\scriptsize\texttt{mart@qti.qualcomm.com}} \\[1.2ex]
\textbf{Paul N. Whatmough} &
\textbf{Babak Ehteshami Bejnordi} \\
Qualcomm AI Research &
NVIDIA \\
{\scriptsize\texttt{pwhatmou@qti.qualcomm.com}} &
{\scriptsize\texttt{babakint@gmail.com}} \\
\end{tabular}
}

\newcommand{\ours}{\textsc{Turnout}}

\iclrfinalcopy
\begin{document}

\maketitle

\begingroup
\renewcommand{\thefootnote}{\fnsymbol{footnote}}
\footnotetext[3]{Qualcomm AI Research is an initiative of Qualcomm Technologies, Inc.} % 
\endgroup
\begin{abstract}

Post-hoc transformer linearization replaces the key-value cache of full attention, which grows linearly with generation length, with a fixed size state.
This change reduces memory consumption but often hurts accuracy.
In reasoning tasks, which require long generations, this effect is especially pronounced.
In this work we introduce \ours, a method that mitigates accuracy loss as a result of linearization, while maintaining its memory-reduction benefits.
We apply post-hoc linearization to a pretrained full-attention model, training a linear path through distillation while leaving the original full-attention path unchanged. 
During response generation, the resulting hybrid model retains both paths and uses a learned policy to control when each layer switches from full to linear attention. 
We train the policy with reinforcement learning using a reward that balances answer correctness and memory use.  
Empirical evaluation of our method shows that the learned adaptive policy achieves a better accuracy-efficiency Pareto frontier on reasoning tasks compared to fixed switching and static hybrid baselines.
\end{abstract}

\input{text/01_introduction}
\input{text/02_related_work}
\input{text/03_method}
\input{text/04_experiments}
\input{text/05_conclusion}

\bibliography{iclr2027_conference}
\bibliographystyle{iclr2027_conference}

\appendix
\input{text/appendix}

\end{document}

%% file: math_commands.tex
\usepackage{amsmath,amsfonts,bm}

\def\eqref#1{equation~\ref{#1}}
\def\1{\bm{1}}

\DeclareMathAlphabet{\mathsfit}{\encodingdefault}{\sfdefault}{m}{sl}
\SetMathAlphabet{\mathsfit}{bold}{\encodingdefault}{\sfdefault}{bx}{n}

%% file: text/01_introduction.tex
\section{Introduction}

Transformer language models remain the dominant architecture for high-quality sequence modeling, but their inference cost grows linearly with context length. During decoding, every generated token attends to an expanding key--value (KV) cache, making long-context generation increasingly memory intensive and limiting throughput once cache movement dominates the runtime~\citep{xiao2023efficient,dao2023flashattention,kuzina2026keygoinglinear}. Linear attention mechanisms offer an attractive alternative: they compress past context into a fixed-size state, so the memory footprint no longer grows linearly with the number of generated tokens~\citep{choromanski2020rethinking,yang2023gated,sun2023retentive,yang2024parallelizing,yang2024gated,team2025kimi}.

Post-hoc linearization is appealing because it starts from an already trained softmax-attention backbone and converts its attention layers into cheaper recurrent alternatives~\citep{mercat2024linearizing,zhang2024lolcats,lan2025liger,kuzina2026keygoinglinear}, instead of training a linear model from scratch. However, directly replacing softmax attention introduces a substantial mismatch with the pretrained model and often reduces accuracy on tasks that rely on precise retrieval or multi-step reasoning~\citep{mercat2024linearizing,zhang2024lolcats,mcdermott2025lola,kuzina2026keygoinglinear}.

This paper focuses on hybrid post-hoc linearization, in which the original multi-head attention (MHA) path remains available alongside a linearized alternative. The two paths share the pretrained backbone and differ only in their attention-specific components. As such, each transformer layer in our model can switch from MHA to linear during generation by generating a linear state from previously generated key and value vectors. We use MHA for the prompt, after which a learned policy decides when each layer switches to linear attention during generation. This matters because the best attention mechanism can depend on the phase of inference. For example, MHA could be preferable during prefill, where the model consumes the prompt and builds a representation of the context, while a linear recurrent path is much cheaper during long decoding, where the key-value (KV) cache would otherwise continue to grow. Jet-Nemotron~\citep{gu2025jet} further shows that the importance of full-attention layers is task-dependent: layers that are critical for MMLU are not necessarily important for math problems and vice versa.

For the linear path, we build on the design of \citet{kuzina2026keygoinglinear}, which reports strong post-hoc linearization results on common reasoning benchmarks and MMLU. We improve their recipe with distillation, as supported by prior work on transforming attention-based models to recurrent or hybrid alternatives~\citep{ro2025fly,wang2024mamba,bick2025llamba}. Furthermore, we combine LoRA adaptation of the attention output projection with short convolutional mixing. As a result, we obtain a linearized model that outperforms all previously reported post-hoc linearized models, as well as pure sliding-window replacements~\citep{jolicoeurmartineau2026slidingwindow}.

We then design a system that can decide, for each layer at each generated token, whether to keep a layer as MHA or switch to linear, simultaneously trying to maximize accuracy while minimizing memory consumption.
Note that, unlike prior work, our method not only decides \emph{which} layers (layer dimension) to switch to linear, but also \emph{when} to do so (token dimension). This flexibility is motivated by the observation that the relative benefits of full and linear attention can vary across tasks and may also depend on the difficulty of individual prompts, making a fixed switching schedule potentially suboptimal.
We formulate this as a reinforcement learning (RL) problem. The policy receives both an accuracy-oriented reward and a memory-aware reward, so it is encouraged to preserve task performance while reducing inference memory. Unlike a fixed manual rule, the learned policy adapts the switch point to each input.Across the five reasoning benchmarks, the learned router scores $82.4\%$ average accuracy, improving over the best manual fixed-switch baseline by $7.8$ percentage points while reducing cache memory by $25.4\%$ relative to full MHA.

Our main contributions are as follows:

\begin{itemize}[leftmargin=*]
  \item We improve post-hoc linearization accuracy on reasoning benchmarks.
  % We use distillation, LoRA adaptation on the output projection, and short convolutional mixing to improve the quality of the linearized path.
  \item We propose \ours{}\footnote{\revised{The name \ours{} refers to a railway turnout, where a train can leave the main track for an alternative route; here, each layer learns when to leave full attention for its linear path.}}, an architecture which enables inference-time switching between MHA and linear attention without training a separate model for each hybrid schedule.
  \item We formulate attention switching as an RL problem with memory-aware and accuracy rewards, and compare learned switching against manual fixed-switch baselines.
\end{itemize}

%% file: text/02_related_work.tex
\section{Related Work}

\paragraph{Linear-time sequence models.} Linear-time sequence models provide the primitives used for post-hoc transformer linearization. Kernelized attention methods such as Performer~\citep{choromanski2020rethinking} and feature-map approaches such as Hedgehog/Porcupine~\citep{zhang2024hedgehog} approximate the softmax attention computation directly. A complementary line develops recurrent or state-based replacements, including Gated Linear Attention (GLA)~\citep{yang2023gated}, Delta Rule~\citep{yang2024parallelizing}, RetNet~\citep{sun2023retentive}, Longhorn~\citep{liu2024longhorn}, Gated Delta Networks (GDN)~\citep{yang2024gated}, Kimi Linear~\citep{team2025kimi}, and transformer--SSM views of Mamba-style models~\citep{dao2024transformers,egorov2025myosotis}. These models show that recurrent state updates can support efficient long-context computation, but they are usually proposed as standalone architectures or pretraining recipes. Our work instead starts from an existing pretrained softmax model and treats linear attention as one path inside a hybrid inference system.

\paragraph{Post-hoc linearization of pretrained transformers.} Post-hoc linearization methods try to close the quality gap introduced by replacing softmax attention in an already pretrained transformer with a cheaper recurrent update. LoLCATs~\citep{zhang2024lolcats} and Liger~\citep{lan2025liger} combine linearized attention with local exact attention such as sliding-window attention, while LoLA~\citep{mcdermott2025lola}, STILL~\citep{meng2026still}, and Lizard~\citep{van2025lizard} preserve selected exact tokens through sparse caches or hybrid token selection. Other approaches improve the conversion through adaptive distillation~\citep{ro2025fly}, direct layerwise linearization~\citep{mercat2024linearizing}, and low-rank adaptation of attention projections~\citep{zhang2024lolcats,mcdermott2025lola}. Most closely related to our work, \citet{kuzina2026keygoinglinear} analyze post-hoc transformer linearization in a frozen-backbone setting and report the strongest results among prior post-hoc linearization methods on common reasoning benchmarks and MMLU. We use their linear attention design as the base linear path in our hybrid model. In addition, we show that for harder reasoning benchmarks, cross-entropy training alone is not sufficient to recover the original model behavior, and we therefore use distillation to improve the linearized path before learning when to switch between MHA and linear attention.

\paragraph{Hybrid models and switching.} Sequence-level hybrid models have shown that attention and recurrent computation can be used at different positions within a sequence. TransMamba~\citep{li2026transmamba} assigns each layer a predefined position at which it switches from attention to Mamba. Oryx~\citep{li2026multimixer} shares most parameters between attention and recurrent mixers and supports switching during prefill and generation, but it pretrains the hybrid model with mixed-mode sequences and selects the mixer schedule manually. These methods establish the feasibility of temporal switching, but their fixed schedules cannot adapt to the input. We instead start from a post-hoc linearized pretrained model and learn an input-dependent switch point for each layer, using fixed switching only as a baseline.

%% file: text/03_method.tex
\section{Linearization}
\label{sec:linearization}

\input{figures/figure_1}

To isolate linearization quality, we first evaluate the linear path without a switching policy. We use Qwen3-8B (thinking model)~\citep{yang2025qwen3technicalreport} and Llama~3.1-8B-Instruct~\citep{grattafiori2024llama3herdmodels} as teachers. Our main evaluation focuses on five reasoning benchmarks: GSM8K (0-shot)~\citep{cobbe2021trainingverifierssolvemath}, MATH-500~\citep{hendrycks2021measuringmathematicalproblemsolving}, GPQA-Diamond~\citep{rein2023gpqagraduatelevelgoogleproofqa}, HumanEval+, and MBPP+~\citep{liu2023codegeneratedchatgptreally}. In this section, we also report two non-reasoning benchmarks, MMLU (5-shot)~\citep{hendrycks2021measuringmassivemultitasklanguage} and LAMBADA (OpenAI)~\citep{paperno2016lambadadatasetwordprediction}, to compare with prior linearization studies that do not report reasoning benchmarks. We omit other common reasoning benchmarks because they are too easy and can be solved by sliding-window attention (SWA) alone.
Our linearization is based on Key to Going Linear (KtGL)~\citep{kuzina2026keygoinglinear}. We improve the performance of their method by using a different training objective, dataset, and longer training.

\paragraph{Linear path and Training Objective}
The linear path combines a Gated Delta Network~\citep{yang2024gated} with a fixed 64-token SWA and 8 sink tokens. We train short convolutions and a rank-128 LoRA adapter on output projection (see Fig.~\ref{fig:linearization-router-overview}a). 
We test three objectives used in prior linearization work: next-token cross-entropy, as used by Liger and KtGL~\citep{lan2025liger,kuzina2026keygoinglinear}; logit distillation, which matches the student's next-token distribution to the teacher's, as used by Mamba in the Llama and Llamba~\citep{wang2024mamba,bick2025llamba}; and activation distillation, which matches intermediate attention outputs, as used in the attention-transfer stage of LoLCATs~\citep{zhang2024lolcats}. 
% We find that a combination of the two distillation losses (logit and activation) works best. 
The objective ablation in Table~\ref{tab:linearization-objective-ablation} (App.~\ref{app:linearization-training}) shows that logit distillation gives most of the improvement on reasoning benchmarks. Combining logit and activation distillation gives the best GSM8K result. The adapter ablation shows that LoRA on $o_{\mathrm{proj}}$ is important for reasoning: rank 128 substantially improves over the no-LoRA variant, while increasing the rank to 256 provides no further gain. We therefore use the mixed distillation objective with rank-128 LoRA. Formal definitions of the losses and the full ablations are provided in App.~\ref{app:linearization-training}.

\paragraph{Training Dataset}
In App.~\ref{app:linearization-data-ablation}, we compare training on DCLM-Edu~\citep{li2024datacomp}, OpenThoughts3 (OT)~\citep{guha2025openthoughtsdatarecipesreasoning}, and DCLM-Edu followed by OT. OT training for 4.5B tokens produces the strongest thinking results among the trained linearized models. We therefore use this setup as the linearized Qwen3 model and apply the same training recipe to Llama.

\paragraph{Sliding-Window Size}
As mentioned above, our linear path combines linear attention with sliding-window attention and sink tokens.
Based on non-reasoning evaluations, recent work has shown that SWA with sink tokens alone can produce strong results, challenging the need for post-hoc linearization~\citep{jolicoeurmartineau2026slidingwindow}.
We verify whether this holds for reasoning benchmarks by comparing a model using only SWA and sink tokens with our linearized model. Notably, we use a sliding-window size of 64 during training and increase the SWA size to 128 and 512 only during evaluation. We keep the number of sink tokens fixed at 8 throughout. 

\begin{figure*}[t]
  \centering
  \includegraphics[width=\textwidth]{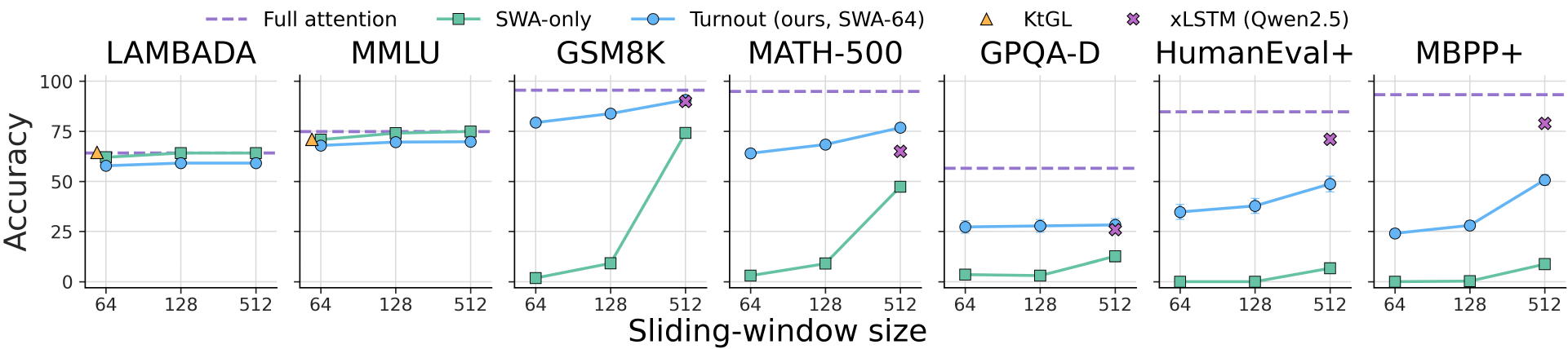}
  % \par\vspace{-1ex}
  {\small\textbf{(a)} Qwen3-8B}
  \par\vspace{0.5ex}
  \includegraphics[width=\textwidth]{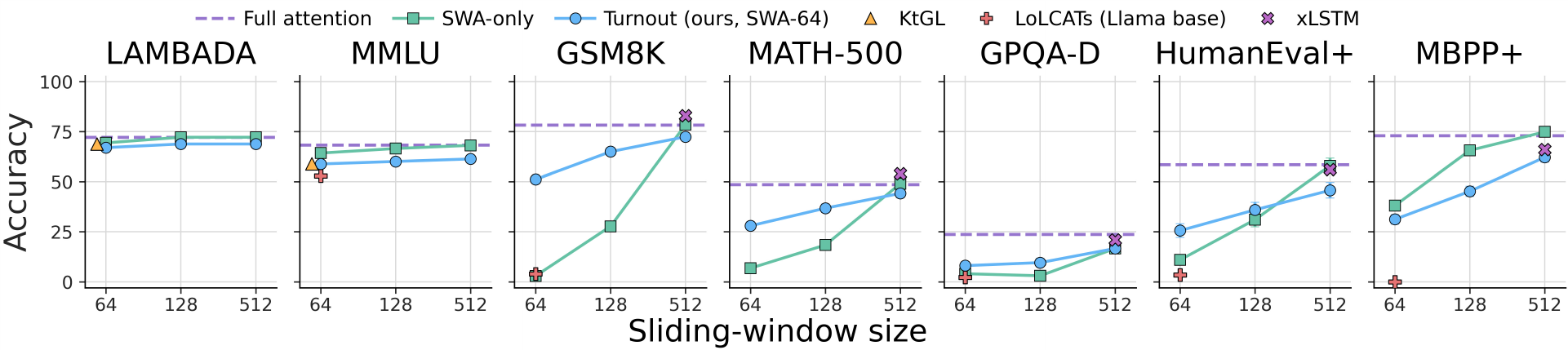}
  % \par\vspace{-1ex}
  {\small\textbf{(b)} Llama~3.1-8B.}
  % \vspace{-10pt}
  \caption{Accuracy across sliding-window sizes for (a) Qwen3-8B and (b) Llama~3.1-8B-Instruct. \ours{} is trained with SWA-64 and evaluated with the three plotted windows. Dashed lines show base model accuracy. KtGL~\citep{kuzina2026keygoinglinear}, LoLCATs~\citep{zhang2024lolcats}, and xLSTM~\citep{beck2026xlstmhybrid} use the values reported in the original papers.}
  \label{fig:swa-window-ablation}
  % \vskip -10pt
\end{figure*}

Figure~\ref{fig:swa-window-ablation} reports accuracy across seven benchmarks.
For both Qwen3-8B and Llama~3.1-8B, increasing the SWA window recovers full-attention performance on LAMBADA and MMLU. Reasoning benchmarks show a different pattern. At SWA-64, training the linear path improves almost every result across the two models. For Qwen3-8B, the linearized model remains stronger on all five reasoning benchmarks even at SWA-512. For Llama~3.1-8B, a 512-token SWA window closes much of the gap and matches full attention on several reasoning benchmarks. The exact values for all window sizes are reported in Appendix~\ref{app:swa-window-ablation}.

%%%%%%%%%%%%%%%%%%%%%%%%%%%%%%%%%%%%%%%%%%%%%%%%%%%%%
% THE ROUTER AND RL PART
%%%%%%%%%%%%%%%%%%%%%%%%%%%%%%%%%%%%%%%%%%%%%%%%%%%%%

\section{Adaptive Hybrid Routing}
\label{sec:routed-hybridization}

\begin{figure*}[t]
  \centering
  \begin{minipage}[t]{0.3\textwidth}
    \centering
    \includegraphics[width=\linewidth]{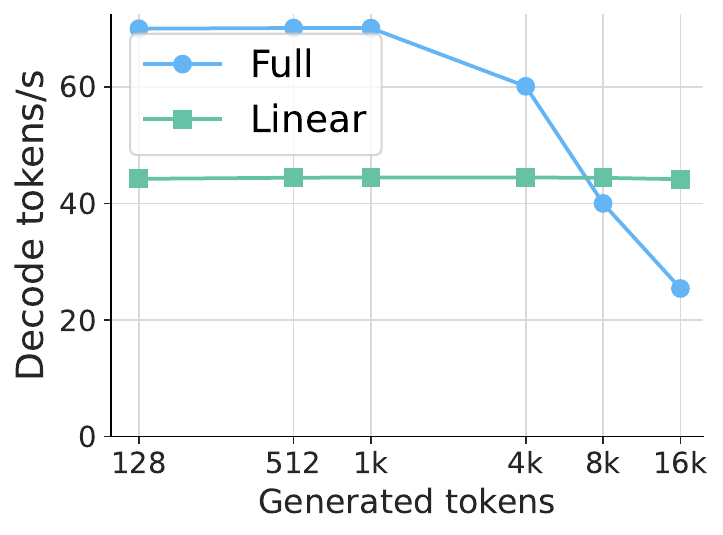}
    \par\smallskip\textbf{(a)} Throughput by length
  \end{minipage}\hfill
  \begin{minipage}[t]{0.3\textwidth}
    \centering
    \includegraphics[width=\linewidth]{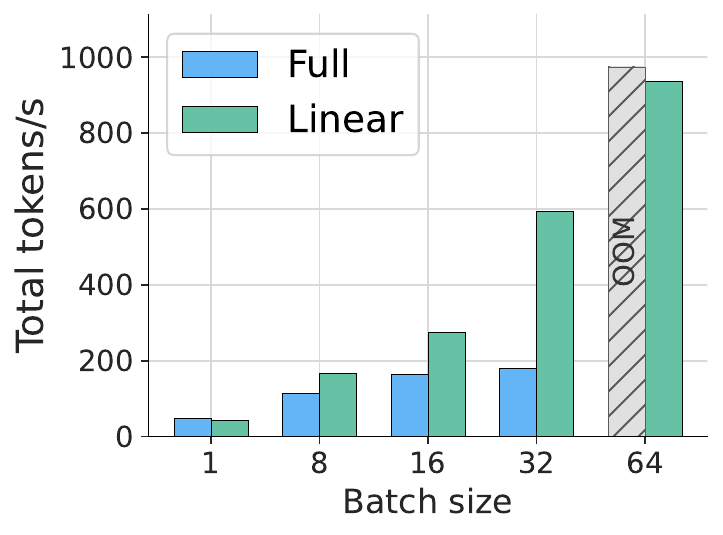}
    \par\smallskip\textbf{(b)} Throughput by batch size
  \end{minipage}\hfill
  \begin{minipage}[t]{0.3\textwidth}
    \centering
    \includegraphics[width=\linewidth]{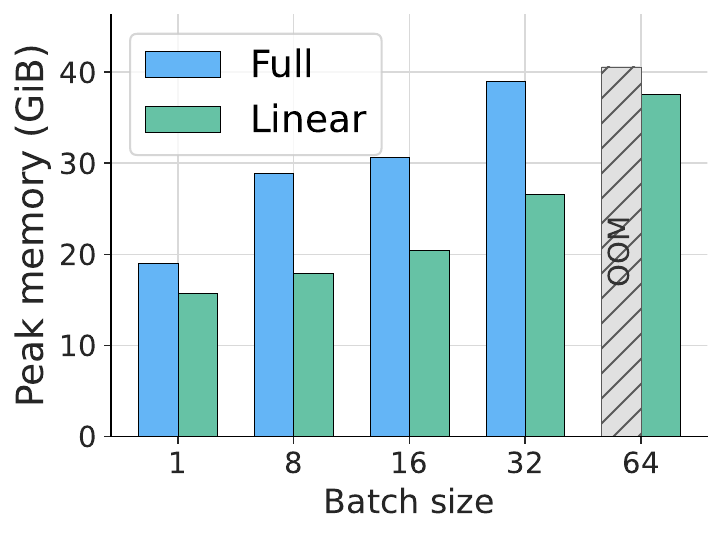}
    \par\smallskip\textbf{(c)} Peak allocated memory
  \end{minipage}
  % \vskip -5pt
  \caption{Inference profiling of the full-attention and linearized Qwen3-8B models on an H100 GPU using the same fixed subset of 128 MATH-500 examples. (a) Throughput across different sequence lengths at batch size 1. (b) Throughput at batch sizes 1, 8, 16, 32, and 64. (c) Peak PyTorch allocated memory at the same batch sizes.}
  \label{fig:profiling-motivation}
  % \vskip -10pt
\end{figure*}

Full attention and linear attention offer complementary advantages during generation.
Full attention is faster for short sequences, whereas linear attention maintains a fixed-size state that is more efficient for long sequences.
To illustrate this trade-off, we profile the full-attention and linearized Qwen3-8B models on the same subset of 128 MATH-500 examples (Fig.~\ref{fig:profiling-motivation}).
At batch size 1, full attention achieves higher token throughput for short sequences, but linear attention becomes faster as the sequence grows (Fig.~\ref{fig:profiling-motivation}a).
Linear attention also uses less peak memory (Fig.~\ref{fig:profiling-motivation}c), allowing larger batches and therefore higher aggregate throughput (Fig.~\ref{fig:profiling-motivation}b).
These complementary regimes motivate \ours{}, which uses full attention at the beginning of generation and switches to linear attention as the response grows longer.

The post-hoc linearization procedure described in the previous section, makes this transition possible. It preserves the original full-attention path and adds a linear path to each layer which shares most of the parameters with MHA. 
We first describe layerwise switching in Sec.~\ref{sec:fixed-switching}, then introduce a lightweight router that decides when each layer should switch (Sec.~\ref{sec:router}) and propose to train it with reinforcement learning (Sec.~\ref{sec:policy-optimization}).
\revised{RL is well suited to this problem because switching decisions interact across layers and tokens, while their effect on answer correctness is known only after generation is complete. It enables an input-dependent router to learn from this delayed, non-differentiable accuracy signal and the efficiency reward.}
Figure~\ref{fig:linearization-router-overview} provides an overview of our proposed method, \ours{}, showing the two attention paths, the router, and the layer-specific switch points.

\subsection{Layerwise Switching}
\label{sec:setting}
\label{sec:fixed-switching}

Let $P$ be the prompt length, $G$ the number of generated tokens, and $T=P+G$. For each layer $\ell\in\{1,\ldots,L\}$, let $s_\ell\in\{0,\ldots,G\}$ be the number of generated tokens processed with full attention and let $\tau_\ell=P+s_\ell$ be the corresponding total-token switch position. The routing mask is
\begin{equation}
  m_{\ell,t}=
  \begin{cases}
    1, & t\leq\tau_\ell,\\
    0, & t>\tau_\ell,
  \end{cases}
\end{equation}
where $m_{\ell,t}=1$ means that layer $\ell$ uses full attention to process token $t$ and $m_{\ell,t}=0$ means that it uses linear attention. Thus $s_\ell=0$ switches immediately after the prompt, while $s_\ell=G$ keeps the layer in full attention for the complete answer.

Full attention stores a KV-cache that grows with the sequence. Linear attention stores a fixed-size recurrent state and the local Sink+SWA state described in Section~\ref{sec:linearization}. 
\revised{Both paths of a layer use the same key and value projections. While a layer runs in full attention, its keys and values are stored in the KV cache as usual, and we also store the two scalar GDN gates of each token. When the layer switches, we compute the GDN state from the cached keys, values, and gates. This state is exactly the state that the linear path would have after processing the same prefix, so switching needs no extra training or tuning. After the switch, we keep only the sink and sliding-window tokens of the KV cache and release the rest from memory (see details in Appendix~\ref{app:switching-implementation}).}

As a deterministic baseline, fixed switching may use the same generated-token budget for every layer and example, i.e.
  $s_\ell=s_{\mathrm{fixed}},
  \text{for all }\ell,$
where the budget $s_{\mathrm{fixed}}$ is selected to satisfy the target KV-cache constraint.

\subsection{Per-Layer Router}
\label{sec:router}

\ours{} replaces the fixed budget with layer- and input-dependent decisions. Before the attention module of each layer that remains in full-attention mode, the router predicts whether that layer should remain in full attention or transition to linear attention. The router combines two signals: a \emph{schedule score} based on the token position and a \emph{context score} based on the hidden representation of the specific prompt. The stay-full logit and probability are
\begin{equation}
  \nu_{\ell,t}=\nu^{\mathrm{schedule}}_{\ell,t}+\nu^{\mathrm{context}}_{\ell,t},
  \qquad
  p^{\mathrm{full}}_{\ell,t}=\sigma(\nu_{\ell,t}).
\end{equation}
where $\sigma$ denotes the sigmoid function.

\paragraph{Schedule Score.}
The schedule score uses position and routing information available before the layer is evaluated. We define the memory break-even point $B=M_{\mathrm{linear}}/c_{\mathrm{KV}}$, where $M_{\mathrm{linear}}$ is the cache memory used by one layer's linear path and $c_{\mathrm{KV}}$ is the memory used by one layer's full-attention KV-cache per token. Thus $B$ is the number of full-attention tokens whose KV-cache uses the same memory as the linear-path cache. It can be computed directly from the memory footprints of the two paths or treated as a tunable reference point that biases the router toward switching after approximately $B$ total tokens. The same break-even point defines the memory reward in Section~\ref{sec:router-training-objective}; we therefore provide the router with position features relative to $B$, aligning its inputs with the efficiency objective used to train it. The schedule input is
\begin{equation}
  \boldsymbol{\phi}_t=
  \left[
    \boldsymbol{\psi}(t);
    \frac{t}{B};
    \operatorname{clip}\!\left(\frac{t}{B},0,1\right);
    \max\!\left(\frac{t-B}{B},0\right);
    \mathbf m_{t-1}
  \right],
  \qquad
  \nu^{\mathrm{schedule}}_{\ell,t}=\operatorname{MLP}_{\ell}(\boldsymbol{\phi}_t),
\end{equation}
where $\boldsymbol{\psi}(t)$ is a log-spaced sinusoidal encoding of the total token position and $\mathbf m_{t-1}$ records which layers were still using full attention at the preceding token. Each layer has an independent small MLP that maps these features to its schedule score.

\paragraph{Context Score.}
The context score enables input-dependent switching decisions. It uses the current pre-attention hidden state $\mathbf h_{\ell,t}$ as a frozen input, so router training does not update the language model. A shared projection produces a compact context representation, followed by a layer-specific linear layer with scalar output:
\begin{equation}
  \mathbf u_{\ell,t}=\operatorname{SiLU}\!\left(W^c\operatorname{sg}[\mathbf h_{\ell,t}]+\mathbf b^c\right),
  \qquad
  \nu^{\mathrm{context}}_{\ell,t}=\operatorname{Linear}_{\ell}(\mathbf u_{\ell,t}).
\end{equation}

Let $\theta$ denote all trainable parameters of the schedule and context branches. The routing policy is the Bernoulli distribution $\pi_\theta=\operatorname{Bernoulli}(p^{\mathrm{full}}_{\ell,t})$. For each layer that is still using full attention, the router samples $m_{\ell,t}\sim\pi_\theta$, with
\begin{equation}
  m_{\ell,t}=1 \;\Rightarrow\; \text{full attention},
  \qquad
  m_{\ell,t}=0 \;\Rightarrow\; \text{switch to linear attention}.
\end{equation}
After the first zero, the mask remains zero and is no longer sampled for that layer. Figure~\ref{fig:linearization-router-overview}b summarizes its inputs and decision. We describe router architecture in more details in App.~\ref{app:router-architecture}.

%%%%%%%%%%%%%%%%%%%%%%%%%%%%%%%%%%%%%%
% Reward
%%%%%%%%%%%%%%%%%%%%%%%%%%%%%%%%%%%%%%%%%%

\subsection{Accuracy and Efficiency Reward}
\label{sec:router-training-objective}

The router is trained to reduce cache memory while preserving answer quality. We use two ways to combine accuracy and efficiency:
\begin{equation}
  R_{\mathrm{add}}
  =a(\hat y,y)+\frac{\alpha}{L}\sum_{\ell=1}^{L}e_\ell,
  \qquad
  R_{\mathrm{scaled}}
  =a(\hat y,y)\left(1+\frac{\alpha}{L}\sum_{\ell=1}^{L}e_\ell\right).
  \label{eq:router-reward}
\end{equation}
Here $\hat y$ is the generated response and $y$ is the reference answer. The accuracy score $a(\hat y,y)$ is one when $\hat y$ matches $y$ and zero otherwise. The score $e_\ell\in[0,1]$ measures the efficiency of layer $\ell$, and $\alpha$ controls the trade-off between accuracy and efficiency. The two rewards differ only for an incorrect answer: $R_{\mathrm{add}}$ keeps the efficiency term, while $R_{\mathrm{scaled}}$ is zero. The latter option does not let the model optimize efficiency at the expense of accuracy. Figure~\ref{fig:router-reward} shows the two choices for $e_\ell$ defined below.

\input{figures/router_reward}We compute $e_\ell$ from the layer's peak cache memory $C_\ell$, measured in equivalent KV-cache tokens. Because the switch position $\tau_\ell$ counts both prompt and generated tokens, a layer that switches at $\tau_\ell$ has a KV cache of $\tau_\ell$ tokens. After the switch, its linear-path cache uses the same memory as $B$ KV-cache tokens. Thus, $C_\ell=\max(\tau_\ell,B)$. If the layer never switches, only the KV-cache is used and $C_\ell=T$. The decay parameter $\Delta>0$ sets the range over which the score falls from one at $C_\ell=B$ to zero at $C_\ell=B+\Delta$.

We use two definitions of $e_\ell$. The \emph{memory-only} score depends only on $C_\ell$. Switching before break-even point $B$ does not reduce peak memory, so these switches have a score of one. For $C_\ell>B$, the score decreases linearly to zero at $C_\ell=B+\Delta$:
\begin{equation}
  e_{\mathrm{mem},\ell}
  =\operatorname{clip}\!\left(\frac{B+\Delta-C_\ell}{\Delta},0,1\right).
  \label{eq:memory-score}
\end{equation}

The memory-only score does not distinguish between switch positions before break-even point $B$. The \emph{break-even-cosine} score gives lower values to early switches. When $P<B$, a switch immediately after the prompt, $\tau_\ell=P$, has a score of zero. The score increases to one at $B$, or at $T$ if the sequence ends before $B$. After $B$, the score decreases smoothly over the same range $\Delta$:
\begin{equation}
  e_{\mathrm{cos},\ell}=
  \begin{cases}
    \cos^{3/2}\!\left(
      \dfrac{\pi}{2}\dfrac{\min(B,T) - \tau_\ell}{\min(B,T)-P}
    \right),
    & C_\ell\leq B \ \text{and}\ P<\min(B,T),\\[6pt]
    \cos^{3/2}\!\left(
      \dfrac{\pi}{2}
      \operatorname{clip}\!\left(\dfrac{C_\ell-B}{\Delta},0,1\right)
    \right),
    & \text{otherwise}.
  \end{cases}
\end{equation}
Subtracting $P$ measures progress after the prompt, and $\min(B,T)$ handles sequences that end before $B$. When $T<B$, never switching has a score of one because the full-attention KV cache remains smaller than the linear-path cache. If $P\geq B$, the first branch is skipped. Just after $B$, the cosine score decreases more slowly than the memory-only score. Both scores are zero at $B+\Delta$.

Each training run chooses a reward form from Equation~\ref{eq:router-reward} and uses the same efficiency-score definition for all layers: $e_\ell=e_{\mathrm{mem},\ell}$ for memory-only or $e_\ell=e_{\mathrm{cos},\ell}$ for break-even cosine.

%%%%%%%%%%%%%%%%%%%%%%%%%%%%%%%%%%%%%%
% GRPO
%%%%%%%%%%%%%%%%%%%%%%%%%%%%%%%%%%%%%%%%%%
\subsection{Router Training: GRPO}
\label{sec:router-training-objective}
\label{sec:policy-optimization}

\begin{wrapfigure}[13]{r}{0.45\textwidth}
\vskip -10pt
  \begin{minipage}{0.45\textwidth}
  \refstepcounter{algorithm}\label{alg:router-training}
  \hrule height .8pt\kern2pt
  \textbf{Algorithm~\thealgorithm}\quad Training the adaptive router\par
  \kern2pt\hrule\kern2pt
  \footnotesize
  \begin{algorithmic}[1]
    \For{prompt--answer pair $(x,y)$}
      \For{$k=1,\ldots,K$}
        \State $\hat y_k,\{m_d^{(k)}\}_{d\in\mathcal D_k}\sim\operatorname{Generate}(x;\theta)$
        \State $g_k\gets\sum_{d\in\mathcal D_k}\nabla_\theta\log\pi_\theta(m_d^{(k)})$
        \State Compute $R_k$ using Equation~\ref{eq:router-reward}
      \EndFor
      \State $A_k\gets\dfrac{R_k-\frac{1}{K}\sum R_j}{\operatorname{std}(R_j)+\delta},\quad k=1,\ldots,K$
      \State $\bar g\gets\dfrac{1}{KLG_{\max}}\sum_{k=1}^{K}A_kg_k$
      \State $\theta\gets\operatorname{OptimizerStep}(\theta,-\bar g)$
    \EndFor
  \end{algorithmic}
  \kern2pt\hrule
  \end{minipage}
\end{wrapfigure}

Algorithm~\ref{alg:router-training} summarizes the training procedure. For rollout $k$, let $\mathcal D_k$ denote the layer-token positions at which a routing-mask value is sampled. This set includes both stay-full values and the final zero for every layer that switches. Let $G_{\max}$ be the maximum generation length.

We use group relative policy optimization (GRPO)~\citep{shao2024deepseekmath} without a learned value model to train the router parameters $\theta$. Both the full-attention and linear-attention paths of the LLM remain frozen, and the hidden states used in the context-aware part of the router are detached.

For each prompt, we sample $K$ responses using different routing decisions and compute their rewards as defined in Section~\ref{sec:router-training-objective}. The advantage of response $k$ is its reward relative to the other responses for the same prompt:
\begin{equation}
  A_k=\frac{R_k-\frac{1}{K}\sum_{j=1}^{K}R_j}{\operatorname{std}(R_j)+\delta},
\end{equation}
where $\delta$ is a small constant for numerical stability. A positive $A_k$ means that the response performed better than the other responses in its group, while a negative value means that it performed worse. \revised{Groups in which all answers are incorrect have zero advantages under the accuracy-scaled reward.}

The router is updated with the log-probability gradient of the sampled routing-mask values:
\begin{equation}
  \nabla_\theta\mathcal L_{\mathrm{router}}
  \propto
  -\frac{1}{KLG_{\max}}
  \sum_{k=1}^{K}A_k
  \sum_{d\in\mathcal D_k}
  \nabla_\theta\log \pi_\theta\left(m_d^{(k)}\right).
\end{equation}
The inner sum includes every sampled mask value, including the zero that switches a layer to linear attention. Thus a positive-advantage response increases the probability of its sampled routing trajectory, while a negative-advantage response decreases it. We use max-completion normalization, dividing by $KLG_{\max}$: the maximum possible number of layer-token decisions in the rollout group. This fixed denominator keeps the overall gradient scale bounded and avoids dividing by the realized number of decisions, which would give each decision in a short or early-switching trajectory greater weight. The normalization does not change $A_k$: every sampled decision has the same coefficient, so trajectories with more routing decisions can still contribute more total gradient. Each group is generated by the current router and used for one update, so the usual GRPO probability ratio is one and clipping does not change the gradient.

%% file: figures/figure_1.tex
\begin{figure*}[t]
  \centering
  \definecolor{FigInk}{HTML}{27313B}
  \definecolor{FigPanel}{HTML}{F5F6F8}
  \definecolor{FigBlue}{HTML}{4774B8}
  \definecolor{FigBlueFill}{HTML}{EAF1FB}
  \definecolor{FigCoral}{HTML}{C86654}
  \definecolor{FigCoralFill}{HTML}{FBEDE9}
  \definecolor{FigTeal}{HTML}{398D83}
  \definecolor{FigTealFill}{HTML}{E8F5F2}
  \definecolor{FigGold}{HTML}{B47B23}
  \definecolor{FigGoldFill}{HTML}{FCF0CF}
  \resizebox{\textwidth}{!}{%
  \begin{tikzpicture}[
    >=Latex,
    line cap=round,
    line join=round,
    flow/.style={-{Latex[length=1.15mm,width=0.72mm]},draw=FigInk!72,line width=0.60pt},
    strongflow/.style={-{Latex[length=1.40mm,width=0.86mm]},draw=FigInk,line width=0.72pt},
    wire/.style={draw=FigInk,line width=0.62pt},
    frozen/.style={draw=FigBlue,fill=FigBlueFill,rounded corners=0.55mm,
                   line width=0.72pt,align=center,inner sep=1.5pt},
    trained/.style={draw=FigCoral,fill=FigCoralFill,rounded corners=0.55mm,
                    line width=0.72pt,align=center,inner sep=1.5pt},
    operation/.style={draw=FigInk!52,fill=white,rounded corners=0.55mm,
                      line width=0.62pt,align=center,inner sep=1.5pt},
    gateop/.style={draw=FigGold,fill=FigGoldFill,rounded corners=0.55mm,
                   line width=0.62pt,align=center,inner sep=1.4pt},
    freezemark/.style={text=FigBlue,font=\scriptsize,inner sep=0pt,anchor=north east},
  ]
    % ------------------------------------------------------------------
    % Panels.
    \fill[FigPanel,rounded corners=1.7mm] (0,0.05) rectangle (9.55,6.02);
    \fill[FigPanel,rounded corners=1.7mm] (9.83,0.05) rectangle (13.35,6.02);
    \fill[FigPanel,rounded corners=1.7mm] (13.63,0.05) rectangle (17.30,6.02);
    \draw[FigInk!13,line width=0.52pt] (9.69,0.25)--(9.69,5.82);
    \draw[FigInk!13,line width=0.52pt] (13.49,0.25)--(13.49,5.82);
    \node[anchor=north west,font=\small\bfseries] at (0.25,5.85)
      {(a) Linearized attention};
    \node[anchor=north west,font=\small\bfseries] at (10.08,5.85)
      {(b) Token routing};
    \node[anchor=north west,font=\small\bfseries] at (13.88,5.85)
      {(c) Adaptive hybrid};

    % ==================================================================
    % (a) Reference-style bottom-up attention block.
    \begin{scope}[xshift=-0.45cm]

    % Input shared by all projections.
    \node[font=\scriptsize] at (3.54,0.20) {$h_\ell$};
    \draw[wire] (3.54,0.33)--(3.54,0.45);
    \draw[wire] (1.05,0.45)--(5.95,0.45);

    % Projection row: beta, alpha, K, V, Q.
    \node[trained,minimum width=0.68cm,minimum height=0.39cm,font=\scriptsize]
      (wb) at (1.05,0.92) {$W_\beta$};
    \node[trained,minimum width=0.68cm,minimum height=0.39cm,font=\scriptsize]
      (wa) at (1.96,0.92) {$W_\alpha$};
    \node[frozen,minimum width=0.68cm,minimum height=0.39cm,font=\scriptsize]
      (wk) at (3.35,0.92) {$W_k$};
    \node[frozen,minimum width=0.68cm,minimum height=0.39cm,font=\scriptsize]
      (wv) at (4.65,0.92) {$W_v$};
    \node[frozen,minimum width=0.68cm,minimum height=0.39cm,font=\scriptsize]
      (wq) at (5.95,0.92) {$W_q$};
    \foreach \n in {wb,wa,wk,wv,wq}
      \draw[strongflow] (\n |- 3.54,0.45)--(\n.south);
    \node[freezemark] at ($(wk.north east)+(-0.035,-0.035)$) {\ding{100}};
    \node[freezemark] at ($(wv.north east)+(-0.035,-0.035)$) {\ding{100}};
    \node[freezemark] at ($(wq.north east)+(-0.035,-0.035)$) {\ding{100}};

    % Alpha and beta gates.
    \node[gateop,minimum width=0.43cm,minimum height=0.35cm,font=\scriptsize]
      (sigb) at (1.05,1.52) {$\sigma$};
    \node[gateop,minimum width=0.43cm,minimum height=0.35cm,font=\scriptsize]
      (siga) at (1.96,1.52) {$\sigma$};
    \draw[strongflow] (wb)--(sigb);
    \draw[strongflow] (wa)--(siga);

    % Frozen Q/K RMSNorm and parameter-free RoPE.
    \node[frozen,minimum width=0.92cm,minimum height=0.44cm,font=\tiny]
      (knorm) at (3.35,1.50) {RMSNorm};
    \node[frozen,minimum width=0.92cm,minimum height=0.44cm,font=\tiny]
      (qnorm) at (5.95,1.50) {RMSNorm};
    \node[operation,minimum width=0.72cm,minimum height=0.36cm,font=\scriptsize]
      (krope) at (3.35,1.99) {RoPE};
    \node[operation,minimum width=0.72cm,minimum height=0.36cm,font=\scriptsize]
      (qrope) at (5.95,1.99) {RoPE};
    \draw[wire] (wk)--(knorm);
    \draw[wire] (wq)--(qnorm);
    \draw[wire] (knorm)--(krope);
    \draw[wire] (qnorm)--(qrope);
    \node[freezemark] at ($(knorm.north east)+(-0.035,-0.035)$) {\ding{100}};
    \node[freezemark] at ($(qnorm.north east)+(-0.035,-0.035)$) {\ding{100}};

    % Pre-convolution Q/K/V fork: straight up to the linear path and right
    % to the exact Sink+SWA path.
    \coordinate (ktap) at (3.35,2.28);
    \coordinate (vtap) at (4.65,2.28);
    \coordinate (qtap) at (5.95,2.28);
    \draw[wire] (krope.north)--(ktap);
    \draw[wire] (wv.north)--(vtap);
    \draw[wire] (qrope.north)--(qtap);
    \draw[wire] (ktap)--(8.63,2.28);
    \foreach \p in {ktap,vtap,qtap}
      \fill[FigInk] (\p) circle (0.040);
    \node[font=\tiny,anchor=south] at (7.25,2.33) {$q,k,v$};

    % Trainable short convolutions.
    \node[trained,minimum width=0.76cm,minimum height=0.37cm,font=\tiny]
      (kconv) at (3.35,2.62) {Conv};
    \node[trained,minimum width=0.76cm,minimum height=0.37cm,font=\tiny]
      (vconv) at (4.65,2.62) {Conv};
    \node[trained,minimum width=0.76cm,minimum height=0.37cm,font=\tiny]
      (qconv) at (5.95,2.62) {Conv};
    \draw[wire] (ktap)--(kconv.south);
    \draw[wire] (vtap)--(vconv.south);
    \draw[wire] (qtap)--(qconv.south);

    % Q/K normalization inside GDN.
    \node[operation,minimum width=0.82cm,minimum height=0.36cm,font=\tiny]
      (kl2) at (3.35,3.11) {$\ell_2$ norm};
    \node[operation,minimum width=0.82cm,minimum height=0.36cm,font=\tiny]
      (ql2) at (5.95,3.11) {$\ell_2$ norm};
    \draw[wire] (kconv)--(kl2);
    \draw[wire] (qconv)--(ql2);

    % Compact replacement modules, aligned like the reference figure.
    \node[trained,minimum width=5.70cm,minimum height=0.76cm] (gdn) at (3.52,3.86) {};
    \node[font=\scriptsize\bfseries] at (3.52,4.02) {GDN};
    \node[font=\scriptsize] (stateprev) at (3.16,3.72) {$S_{t-1}$};
    \node[font=\scriptsize] (statenext) at (4.05,3.72) {$S_t$};
    \draw[strongflow] (stateprev)--(statenext);
    \draw[strongflow] (sigb.north)--(sigb.north |- gdn.south);
    \draw[strongflow] (siga.north)--(siga.north |- gdn.south);
    \draw[strongflow] (kl2.north)--(kl2.north |- gdn.south);
    \draw[strongflow] (vconv.north)--(vconv.north |- gdn.south);
    \draw[strongflow] (ql2.north)--(ql2.north |- gdn.south);
    \node[font=\tiny,anchor=west] at (1.13,3.25) {$\beta_t$};
    \node[font=\tiny,anchor=west] at (2.04,3.25) {$\alpha_t$};

    \node[operation,minimum width=2.38cm,minimum height=0.90cm] (swa) at (8.63,3.86) {};
    \node[font=\scriptsize\bfseries] at (8.63,4.06) {Sink + SWA};
    \node[draw=FigGold,fill=FigGoldFill,rounded corners=0.20mm,
          minimum width=0.19cm,minimum height=0.27cm,inner sep=0pt] at (7.91,3.72) {};
    \node[draw=FigGold,fill=FigGoldFill,rounded corners=0.20mm,
          minimum width=0.19cm,minimum height=0.27cm,inner sep=0pt] at (8.17,3.72) {};
    \node[font=\tiny,text=FigInk!38] at (8.42,3.72) {$\cdots$};
    \foreach \xx in {8.68,8.94,9.20}
      \node[draw=FigTeal,fill=FigTealFill,rounded corners=0.20mm,
            minimum width=0.19cm,minimum height=0.27cm,inner sep=0pt] at (\xx,3.72) {};
    \draw[strongflow] (8.63,2.28)--(swa.south);

    % Add the two paths.
    \node[circle,draw=FigInk!52,fill=white,minimum size=0.37cm,
          inner sep=0pt,font=\scriptsize\bfseries] (mix) at (6.90,3.86) {$+$};
    \draw[strongflow] (gdn.east)--(mix.west);
    \draw[strongflow] (swa.west)--(mix.east);

    % Frozen output projection plus trainable LoRA, shown as parallel terms.
    \node[draw=FigInk!38,fill=none,rounded corners=0.75mm,line width=0.58pt,
          minimum width=2.55cm,minimum height=0.72cm] (outblock) at (6.90,5.25) {};
    \node[frozen,minimum width=0.84cm,minimum height=0.39cm,font=\scriptsize]
      (wo) at (6.18,5.25) {$W_o$};
    \node[font=\scriptsize\bfseries] at (6.90,5.25) {$+$};
    \node[trained,minimum width=0.74cm,minimum height=0.39cm,font=\tiny]
      (lora) at (7.62,5.25) {LoRA};
    \draw[strongflow] (mix.north)--(outblock.south);
    \node[freezemark] at ($(wo.north east)+(-0.035,-0.035)$) {\ding{100}};
    \draw[strongflow] (outblock.north)--++(0,0.27)
      node[above,font=\scriptsize] {$h_{\ell+1}$};
    \end{scope}

    % ==================================================================
    % (b) Bottom-to-top token-level router schematic.
    \begin{scope}[xshift=-0.35cm]

    % Three router inputs along the bottom.
    \fill[FigInk!30] (11.15,0.86) rectangle (11.39,0.93);
    \fill[FigInk!42] (11.15,0.74) rectangle (11.51,0.81);
    \fill[FigInk!35] (11.15,0.62) rectangle (11.45,0.69);
    \node[font=\tiny] at (11.32,0.39) {$h^{\mathrm{pre}}_{\ell,t}$};
    \coordinate (hin) at (11.32,1.04);

    \draw[FigInk!48,line width=0.56pt] (11.72,0.77)--(12.24,0.77);
    \foreach \xx in {11.79,11.98,12.17}
      \draw[FigInk!48,line width=0.56pt] (\xx,0.71)--(\xx,0.83);
    \fill[FigCoral] (11.98,0.77) circle (0.047);
    \node[font=\tiny] at (11.98,0.39) {$\phi_t$};
    \coordinate (tin) at (11.98,1.04);

    \node[draw=FigBlue,fill=FigBlueFill,rounded corners=0.18mm,
          minimum width=0.16cm,minimum height=0.23cm,inner sep=0pt] at (12.57,0.77) {};
    \node[draw=FigBlue,fill=FigBlueFill,rounded corners=0.18mm,
          minimum width=0.16cm,minimum height=0.23cm,inner sep=0pt] at (12.78,0.77) {};
    \node[draw=FigTeal,fill=FigTealFill,rounded corners=0.18mm,
          minimum width=0.16cm,minimum height=0.23cm,inner sep=0pt] at (12.99,0.77) {};
    \node[font=\tiny] at (12.78,0.39) {$\mathbf m_{t-1}$};
    \coordinate (min) at (12.78,1.04);

    % One compact router per layer.
    \node[trained,minimum width=2.22cm,minimum height=0.72cm,
          font=\scriptsize\bfseries] (router) at (11.98,2.08)
      {Router$_\ell$\; $\pi_{\theta_\ell}$};
    \coordinate (rleft) at ($(router.south)+(-0.66,0)$);
    \coordinate (rmid) at (router.south);
    \coordinate (rright) at ($(router.south)+(0.80,0)$);
    \draw[strongflow] (hin)--(rleft);
    \draw[strongflow] (tin)--(rmid);
    \draw[strongflow] (min)--(rright);

    \node[circle,draw=FigInk!54,fill=white,minimum size=0.52cm,
          inner sep=0pt,font=\tiny] (choice) at (11.98,3.05) {$a_{\ell,t}$};
    \draw[strongflow] (router.north)--(choice.south);

    \node[frozen,minimum width=1.28cm,minimum height=0.92cm] (full) at (11.18,4.25) {};
    \node[font=\scriptsize\bfseries] at (11.18,4.47) {Full};
    \draw[FigBlue!58,fill=FigBlueFill,rounded corners=0.24mm,line width=0.48pt]
      (10.88,3.91) rectangle (11.19,4.17);
    \fill[FigBlue!70] (10.92,3.95) rectangle (10.99,4.02);
    \fill[FigBlue!55] (10.92,4.06) rectangle (10.99,4.13);
    \fill[FigBlue!55] (11.03,3.95) rectangle (11.10,4.02);
    \foreach \xx in {11.34,11.50}
      \node[draw=FigBlue!68,fill=white,minimum width=0.10cm,
            minimum height=0.16cm,inner sep=0pt] at (\xx,4.03) {};

    \node[draw=FigTeal,fill=FigTealFill,rounded corners=0.55mm,
          line width=0.72pt,minimum width=1.28cm,minimum height=0.92cm]
      (linear) at (12.78,4.25) {};
    \node[font=\scriptsize\bfseries] at (12.78,4.47) {Linear};
    \node[font=\scriptsize] at (12.48,4.04) {$S_t$};
    \draw[FigTeal,-{Latex[length=1.0mm,width=0.60mm]},line width=0.48pt]
      (12.36,3.97) arc[start angle=210,end angle=-105,radius=0.20];
    \node[draw=FigGold,fill=FigGoldFill,minimum width=0.11cm,
          minimum height=0.18cm,inner sep=0pt] at (12.87,4.02) {};
    \node[draw=FigTeal,fill=white,minimum width=0.11cm,
          minimum height=0.18cm,inner sep=0pt] at (13.05,4.02) {};

    \draw[-{Latex[length=1.40mm,width=0.86mm]},draw=FigBlue,line width=0.72pt]
      (choice.north west)--(full.south);
    \draw[-{Latex[length=1.40mm,width=0.86mm]},draw=FigTeal,line width=0.72pt]
      (choice.north east)--(linear.south);
    \end{scope}

    % ==================================================================
    % (c) The token sequence moves through the model depth: tokens, layer,
    % updated tokens, next layer.  Each layer exposes the two routed modules.
    \foreach \x/\lab in {14.72/1,15.11/2,15.50/3,15.89/4,16.28/5}
      \node[font=\tiny] at (\x,4.98) {$t_{\lab}$};

    % Four layer stages placed between successive token sequences.
    \foreach \name/\y/\lab in {one/1.05/1,two/2.05/2,three/3.05/3,four/4.05/4}{
      \node[operation,fill=white,minimum width=2.40cm,minimum height=0.38cm]
        (layer\name) at (15.50,\y) {};
      \node[font=\scriptsize\bfseries] at (14.52,\y) {$\ell_{\lab}$};
      \node[frozen,minimum width=0.66cm,minimum height=0.25cm,
            inner sep=0.6pt,font=\tiny] at (15.18,\y) {Full};
      \node[draw=FigTeal,fill=FigTealFill,rounded corners=0.45mm,
            line width=0.66pt,minimum width=0.70cm,minimum height=0.25cm,
            inner sep=0.6pt,font=\tiny] at (15.98,\y) {Linear};
    }

    % Input token sequence.
    \foreach \x in {14.72,15.11,15.50,15.89,16.28}
      \node[draw=FigInk!38,fill=white,rounded corners=0.20mm,
            minimum width=0.20cm,minimum height=0.22cm,inner sep=0pt] at (\x,0.55) {};

    % Tokens after layer 1.
    \foreach \x in {14.72}
      \node[draw=FigBlue,fill=FigBlueFill,rounded corners=0.20mm,
            minimum width=0.20cm,minimum height=0.22cm,inner sep=0pt] at (\x,1.55) {};
    \foreach \x in {15.11,15.50,15.89,16.28}
      \node[draw=FigTeal,fill=FigTealFill,rounded corners=0.20mm,
            minimum width=0.20cm,minimum height=0.22cm,inner sep=0pt] at (\x,1.55) {};

    % Tokens after layer 2.
    \foreach \x in {14.72,15.11,15.50}
      \node[draw=FigBlue,fill=FigBlueFill,rounded corners=0.20mm,
            minimum width=0.20cm,minimum height=0.22cm,inner sep=0pt] at (\x,2.55) {};
    \foreach \x in {15.89,16.28}
      \node[draw=FigTeal,fill=FigTealFill,rounded corners=0.20mm,
            minimum width=0.20cm,minimum height=0.22cm,inner sep=0pt] at (\x,2.55) {};

    % Tokens after layer 3.
    \foreach \x in {14.72,15.11}
      \node[draw=FigBlue,fill=FigBlueFill,rounded corners=0.20mm,
            minimum width=0.20cm,minimum height=0.22cm,inner sep=0pt] at (\x,3.55) {};
    \foreach \x in {15.50,15.89,16.28}
      \node[draw=FigTeal,fill=FigTealFill,rounded corners=0.20mm,
            minimum width=0.20cm,minimum height=0.22cm,inner sep=0pt] at (\x,3.55) {};

    % Tokens after layer 4.
    \foreach \x in {14.72,15.11,15.50,15.89}
      \node[draw=FigBlue,fill=FigBlueFill,rounded corners=0.20mm,
            minimum width=0.20cm,minimum height=0.22cm,inner sep=0pt] at (\x,4.55) {};
    \foreach \x in {16.28}
      \node[draw=FigTeal,fill=FigTealFill,rounded corners=0.20mm,
            minimum width=0.20cm,minimum height=0.22cm,inner sep=0pt] at (\x,4.55) {};

    % The sequence itself flows upward through every layer.
    \draw[strongflow] (15.50,0.68)--(layerone.south);
    \draw[draw=FigInk,line width=0.72pt] (layerone.north)--(15.50,1.42);
    \draw[draw=FigInk,line width=0.72pt] (15.50,1.68)--(layertwo.south);
    \draw[draw=FigInk,line width=0.72pt] (layertwo.north)--(15.50,2.42);
    \draw[draw=FigInk,line width=0.72pt] (15.50,2.68)--(layerthree.south);
    \draw[draw=FigInk,line width=0.72pt] (layerthree.north)--(15.50,3.42);
    \draw[draw=FigInk,line width=0.72pt] (15.50,3.68)--(layerfour.south);
    \draw[strongflow] (layerfour.north)--(15.50,4.42);
  \end{tikzpicture}%
  }
  % \vskip -10pt
  \caption{Overview of \ours{}: linearized attention, token-level routing, and adaptive layerwise switching. \textbf{(a)} Frozen pretrained modules
  (blue snowflakes) are retained. Trainable gates, short convolutions, GDN, and
  LoRA form the linearized branch; exact Sink+SWA runs in parallel.
  \textbf{(b)} The router at layer $\ell$ uses the pre-attention hidden state,
  position and break-even features, and the preceding token's routing mask to
  select exactly one attention path. \textbf{(c)} Across a token sequence,
  different layers can switch from full attention (blue) to linear attention
  (teal) at different times.}
  % \vskip -10pt
  \label{fig:linearization-router-overview}
\end{figure*}

%% file: figures/router_reward.tex
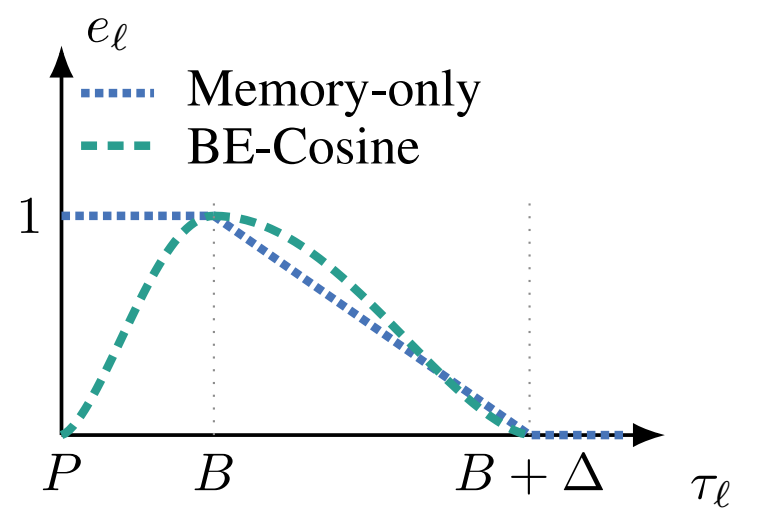
\begin{wrapfigure}{r}{0.36\columnwidth}
  \vspace{-10pt}
  \centering
  \definecolor{RewardBlue}{HTML}{4774B8}
  \definecolor{RewardGreen}{HTML}{2A9D8F}
  \begin{tikzpicture}[x=0.72cm,y=1.45cm,>=Latex]
    \draw[->,line width=0.7pt] (0,0) -- (5.55,0);
    \draw[->,line width=0.7pt] (0,0) -- (0,1.78);
    \node[anchor=north west,font=\normalsize] at (5.60,-0.10) {$\tau_\ell$};
    \node[above=2pt,right=1pt,font=\normalsize] at (0,1.78) {$e_\ell$};
    \draw[RewardBlue,line width=1.6pt,densely dotted] (0,1) -- (1.4,1) -- (4.3,0) -- (5.2,0);
    \draw[RewardGreen,line width=1.6pt,densely dashed]
      plot[domain=0:1.4,samples=40] (\x,{(sin(90*\x/1.4))^1.5})
      plot[domain=1.4:4.3,samples=40] (\x,{(cos(90*(\x-1.4)/2.9))^1.5});
    \draw[dotted,draw=black!45] (1.4,0) -- (1.4,1.08);
    \draw[dotted,draw=black!45] (4.3,0) -- (4.3,1.08);
    \node[below,font=\normalsize] at (0,0) {$P$};
    \node[below,font=\normalsize] at (1.4,0) {$B$};
    \node[below,font=\normalsize] at (4.3,0) {$B+\Delta$};
    \node[left,font=\normalsize] at (0,1) {$1$};
    \draw[RewardBlue,line width=1.6pt,densely dotted] (0.18,1.56) -- (0.85,1.56);
    \node[anchor=west,font=\normalsize] at (0.98,1.56) {Memory-only};
    \draw[RewardGreen,line width=1.6pt,densely dashed] (0.18,1.32) -- (0.85,1.32);
    \node[anchor=west,font=\normalsize] at (0.98,1.32) {BE-Cosine};
  \end{tikzpicture}
  \vspace{-15pt}
  \caption{Per-layer efficiency scores when $P<B$. A layer with $\tau_\ell=P$ switches immediately after the prompt. The memory-only score is one for switch positions between $P$ and $B$. The Break-even-cosine score increases from zero at $P$ to one at $B$. Both scores are zero at $B+\Delta$.}
  \label{fig:router-reward}
  % \vspace{-12pt}
\end{wrapfigure}

%% file: text/04_experiments.tex
% \newpage
\section{Experiments} \label{sec:hybrid-experiments}

\subsection{Experimental Setup}
All routing experiments use the Qwen3-8B full-attention model and its SWA-64+GDN linear path. The linear path was trained on 4.5B OpenThoughts3 tokens, as described in Section~\ref{sec:linearization} and Table~\ref{tab:linearization-data-ablation}. In these experiments we only train the router parameters using Algorithm~\ref{alg:router-training}, while the full-attention and linear-path parameters remain frozen.

The router training data is drawn from the GuRU~\citep{cheng2025revisiting} dataset. We combine its math and code subsets and filter this mixture to retain only examples that the full Qwen3-8B model answers correctly, ensuring that router training focuses on preserving existing model capabilities while improving memory efficiency. Each router is trained for 100 steps on four GPUs, using a per-GPU batch size of 8 and a group size of $K=4$. Unless specified otherwise, we set the efficiency weight $\alpha$ to $0.1$, the break-even point $B$ to $512$ tokens, and the decay parameter $\Delta$ to $8192$ tokens. We do not tune the optimizer hyperparameters and keep them fixed across experiments. Appendix~\ref{app:router-training-hyperparameters} and Table~\ref{tab:router-training-hyperparameters} provide the complete router-training configuration. Evaluation uses the benchmark protocols described in Section~\ref{sec:linearization}.

\subsection{Accuracy-Memory Trade-off}

\begin{figure*}[t]
  \centering
  \includegraphics[width=\textwidth]{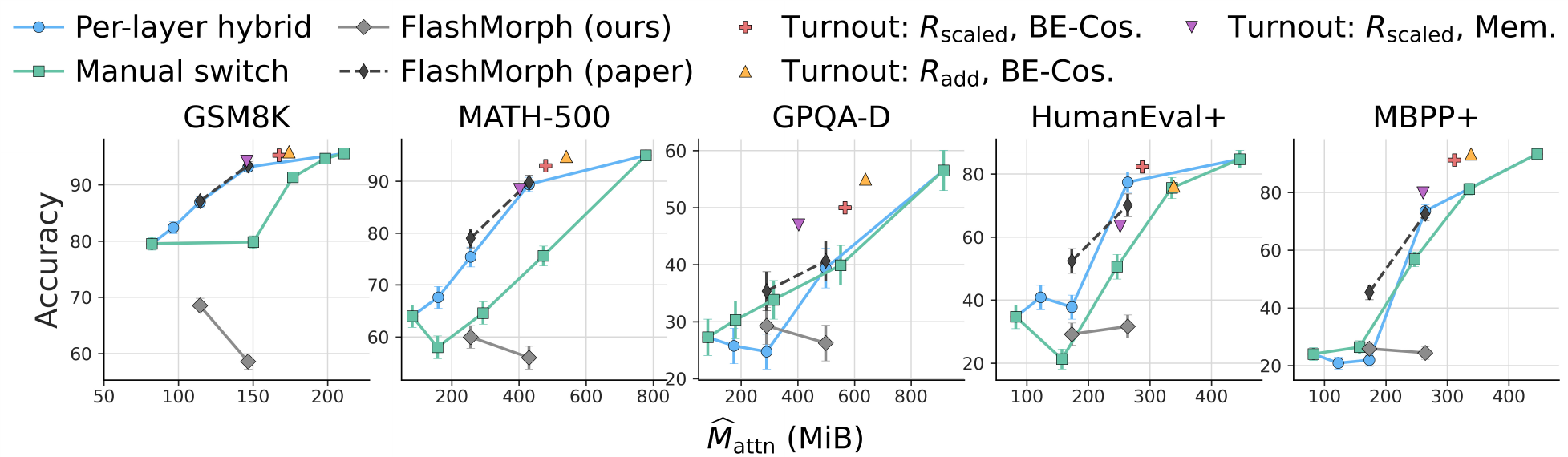}
\vskip -15pt
  \caption{Accuracy against estimated cache memory for Qwen3-8B. The curves
  show per-layer hybridization, \revised{FlashMorph~\citep{lan2026morphing},} and manual switching. The separate markers show
  \ours{} routers trained with different reward functions.}
\vskip -5pt
  \label{fig:hybridization-memory-accuracy}
\end{figure*}

We compare different rewards defined in Section~\ref{sec:router-training-objective}. We use accuracy on reasoning benchmarks to measure the quality of the learned router policy and a cache-memory proxy as the efficiency metric. We calculate the total cache size at each decoding step and use the maximum across all decoding steps as the memory proxy for a given response.
For unswitched layers, we assume that key-value states are stored in bfloat16. Switched layers store the fixed linear-attention state in float32 and the sliding-window cache and attention sinks in bfloat16. We then average this proxy over all generated traces in a given benchmark.

As upper and lower bounds, we report the performance of the all-full and all-linear models. We also compare the learned router with two fixed strategies: per-layer hybrid models, in which we keep $10\%$, $25\%$, or $50\%$ of the layers in full-attention mode, and manual switching, in which we manually set the switching point to 1k, 2k, or 4k tokens.
For each per-layer hybrid model, we select the layers with the largest final activation-distillation losses and use the trained linear path in all other layers. The full-attention layers use the original Qwen3-8B weights, whereas the linear layers use the weights from the all-linear model. We do not retrain these hybrid models.
\revised{We also compare with FlashMorph~\citep{lan2026morphing}, which builds the same per-layer hybrid models but selects the full-attention layers with trained per-layer gates. We use two layer rankings: FlashMorph (ours) uses the ranking from our reimplementation, and FlashMorph (paper) uses the ranking that \citet{lan2026morphing} report for Qwen3-8B. Appendix~\ref{app:flashmorph} gives the details and the selected layers.}
For manual switching, all layers use full attention for the specified number of generated tokens and then switch to the linear path.

In Figure~\ref{fig:hybridization-memory-accuracy}, we plot the resulting \ours{} models and the baselines on the accuracy-memory plane. Early in training, we observed that the additive reward with memory-only efficiency produced a collapsed router that always switched to linear attention immediately. Therefore, we excluded this configuration from the experiment sweep. Figure~\ref{fig:hybridization-memory-accuracy} shows that learned switching improves the accuracy–memory trade-off over manual switching and static per-layer hybridization, especially on GPQA-D and MBPP+. \revised{With the published ranking, FlashMorph is more accurate than our per-layer selection at $25\%$ full layers and close to it at $50\%$. However, \ours{} router is still more accurate than both FlashMorph variants at comparable cache memory on all five benchmarks.} These results suggest that learning which layers retain full attention, and for how long, improves accuracy at comparable cache memory. We report detailed results in Table~\ref{tab:accuracy-memory-results}.

\paragraph{Router Analysis}

\begin{figure}[t]
  \centering
  \includegraphics[width=\textwidth]{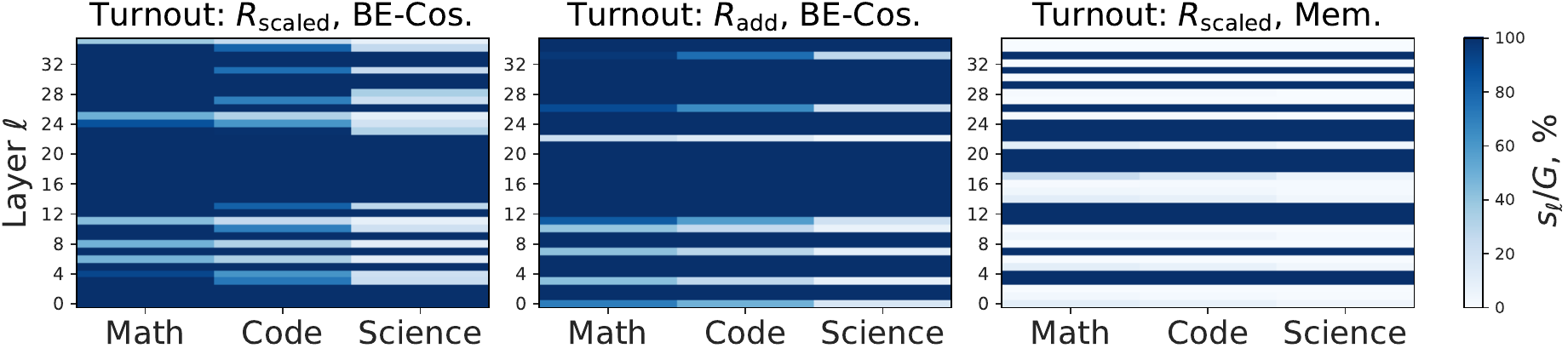}
  \vskip -5pt
  \caption{Layer-wise routing strategies for the cosine-scaled, cosine-additive, and memory-scaled routers. Benchmarks are grouped into math, code, and science. Each cell shows the median fraction of the generated response processed with full attention. Darker cells denote later switching.}
  \vskip -5pt
  \label{fig:router-layer-profile}
\end{figure}

Figure~\ref{fig:router-layer-profile} compares router learned with three different rewards. The scaled break-even-cosine router shows the strongest task dependence, switching several layers earlier for code and science than for math. The additive variant keeps more layers in full attention, whereas the memory-only variant learns an almost task-independent pattern in which each layer switches either very early or very late. Thus, the reward design strongly affects the routing strategy learned by \ours{}.

\paragraph{Ablations}
\begin{table*}[t]
  \centering
  \footnotesize
  \caption{\ours{} ablations of the reward hyperparameters and context input \revised{on Qwen3-8B, and \ours{} on Llama~3.1-8B-Instruct}. }
  \label{tab:reward-ablation}
  \resizebox{\textwidth}{!}{%
  \begin{tabular}{@{}ccccccccccc@{}}
    \toprule
    \multicolumn{4}{c}{Router config.} & \multicolumn{6}{c}{Accuracy} & Cache saving \\
    \cmidrule(lr){1-4}\cmidrule(lr){5-10}\cmidrule(l){11-11}
    \revised{Score} & Context & $\Delta$ & $\alpha$ & GSM8K & GPQA-D & MATH-500 & MBPP+ & HE+ & Avg. & (\%) $\uparrow$ \\
    \midrule
    \multicolumn{11}{l}{\revised{\textit{Qwen3-8B}}} \\
    \multicolumn{4}{l}{Full attention} & 95.6 & 56.6 & 95.0 & 93.4 & 84.8 & 85.1 & 0.0 \\
    \multicolumn{4}{l}{All linear} & 79.5 & 27.3 & 64.0 & 24.1 & 34.8 & 45.9 & 69.4 \\
    \midrule
    \revised{BE-Cos.} & \ding{55} & 8K & 0.1 & 94.5 & 45.5 & 91.4 & 77.0 & \textbf{82.9} & 78.3 & 17.4 \\
    \revised{BE-Cos.} & $\checkmark$ & 8K & 0.1 & 95.3 & 50.0 & 93.0 & 91.3 & 82.3 & 82.4 & \textbf{25.4} \\
    \revised{BE-Cos.} & $\checkmark$ & 16K & 0.1 & \textbf{95.6} & 54.0 & \textbf{94.6} & \textbf{93.7} & 79.9 & \textbf{83.6} & 10.3 \\
    % $\checkmark$ & 4K & 0.1 & 93.9 & 47.5 & -- & 62.3 & 72.0 & -- & -- \\
    \revised{BE-Cos.} & $\checkmark$ & 8K & 1.0 & 94.7 & 46.0 & 93.0 & 68.0 & 76.8 & 75.7 & 14.6 \\
    \revised{BE-Cos.} & $\checkmark$ & 16K & 1.0 & 94.9 & \textbf{59.6} & 93.8 & 91.5 & 76.2 & 83.2 & 14.7 \\
    \specialrule{1pt}{2pt}{2pt}
    \multicolumn{11}{l}{\revised{\textit{Llama 3.1-8B}}} \\
    \multicolumn{4}{l}{\revised{Full attention}} & \revised{78.2} & \revised{23.7} & \revised{48.6} & \revised{73.0} & \revised{58.5} & \revised{56.4} & \revised{0.0} \\
    \multicolumn{4}{l}{\revised{All linear}} & \revised{51.2} & \revised{8.1} & \revised{28.0} & \revised{31.2} & \revised{25.6} & \revised{28.8} & \revised{$-68.4$} \\
    \midrule
    \revised{Mem.} & \revised{$\checkmark$} & \revised{1K} & \revised{1.0} & \revised{78.1} & \revised{23.2} & \revised{49.2} & \revised{71.4} & \revised{57.9} & \revised{56.0} & \revised{$-1.4$} \\
    \revised{BE-Cos.} & \revised{$\checkmark$} & \revised{1K} & \revised{1.0} & \revised{77.9} & \revised{26.0} & \revised{52.0} & \revised{71.7} & \revised{60.4} & \revised{57.6} & \revised{1.3} \\
    \bottomrule
  \end{tabular}%
  }
\end{table*}

To study how the reward hyperparameters affect the learned policy, we vary the decay parameter $\Delta$ and the efficiency weight $\alpha$ while fixing the objective to the accuracy-scaled reward with the break-even-cosine efficiency score. We also ablate the context branch by training a schedule-only router that does not receive the hidden-state context input.

Table~\ref{tab:reward-ablation} reports benchmark accuracy and relative cache savings for each configuration. For each sample, we compare the router's peak cache-memory proxy $M_{\mathrm{router}}$ with the full-attention KV-cache $M_{\mathrm{full}}$ and report the relative saving $1-M_{\mathrm{router}}/M_{\mathrm{full}}$. \revised{The full-attention reference uses the same prompt and response length as each routed sample, to ensure that the savings do not come from shorter responses. \ours{} and full attention also generate responses of similar length on all five benchmarks (Appendix~\ref{app:routed-lengths}).} The context input improves the accuracy-memory trade-off: it raises average accuracy from $78.3$ to $82.4$ while increasing cache saving from $17.4\%$ to $25.4\%$ given all other hyperparameter fixed. Increasing $\Delta$ to $16\mathrm{K}$ further raises average accuracy to $83.6$, only $1.5$ points below full attention, but reduces cache saving to $10.3\%$.

\revised{\paragraph{Llama results} Table~\ref{tab:reward-ablation} also reports \ours{} on Llama~3.1-8B-Instruct. Both routers keep the accuracy of full attention ($56.0$ and $57.6$ average accuracy against $56.4$), while the all-linear model loses $27.6$ points. The cache saving is close to zero because Llama~3.1-8B-Instruct does not generate long reasoning traces. One linear layer uses the memory of $584$ KV-cache tokens ($512$ for the GDN state, $64$ for the sliding window and $8$ for the sink tokens). For about $97\%$ of the GSM8K and code samples, prompt and response together have fewer than $584$ tokens, so full attention uses less memory than the linear path, and the all-linear model has a negative saving. With BE-Cos., this effect is mitigated as intended by the break-even point $B$.}

\begin{figure}[t]
  \centering
  \begin{minipage}[t]{0.3\textwidth}
    \centering
    \includegraphics[width=\linewidth]{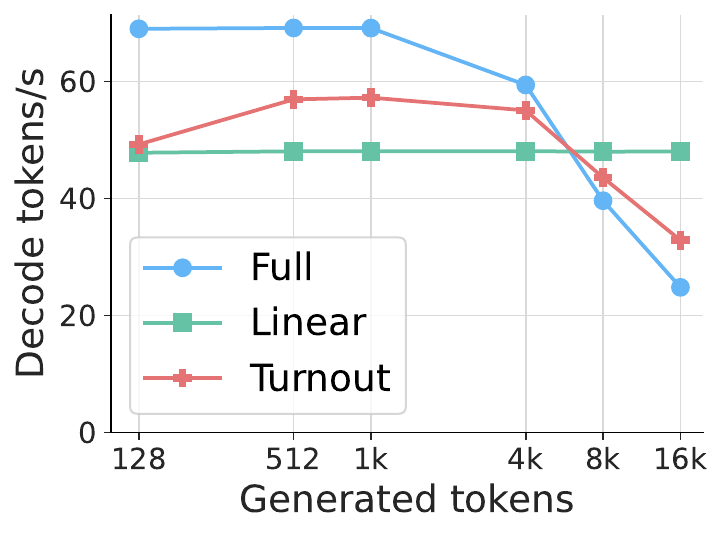}
    \par{\revisioncolor\footnotesize\textbf{(a)} Throughput}
  \end{minipage}\hfill
  \begin{minipage}[t]{0.3\textwidth}
    \centering
    \includegraphics[width=\linewidth]{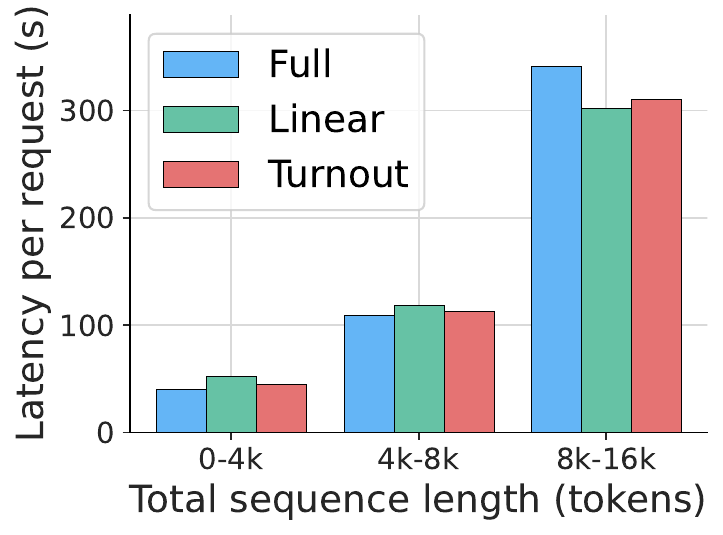}
    \par{\revisioncolor\footnotesize\textbf{(b)} Latency per problem}
  \end{minipage}\hfill
  \begin{minipage}[t]{0.3\textwidth}
    \centering
    \includegraphics[width=\linewidth]{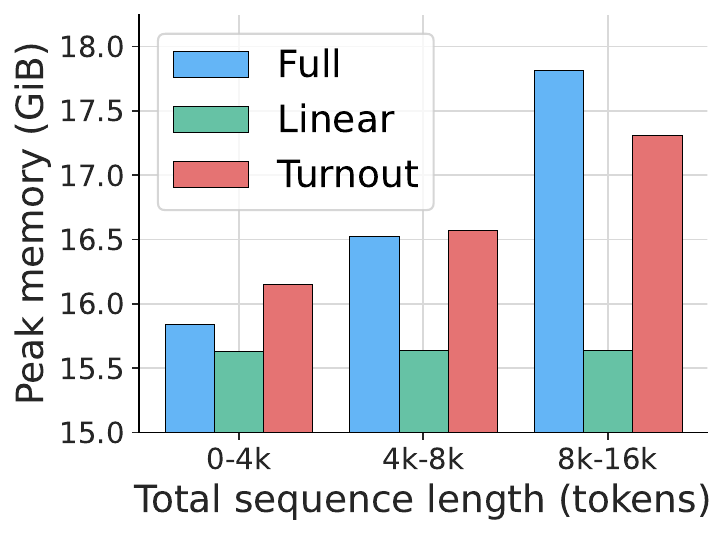}
    \par{\revisioncolor\footnotesize\textbf{(c)} Peak memory}
  \end{minipage}
  \vskip -3pt
  {\revisioncolor
  \caption{Throughput and memory footprint measured on MATH500 using a signle H100 GPU.}
  \label{fig:measured-cost}
  }
  \vskip -15pt
\end{figure}

\subsection{Memory Footprint and Parameter Overhead}

\begin{wraptable}[11]{r}{0.52\textwidth}
  \vspace{-25pt}
  \centering
  \footnotesize
  \setlength{\tabcolsep}{3pt}
  \caption{Memory and parameter overhead.}
  \label{tab:memory-overhead}
  \begin{tabular}{@{}lccccc@{}}
    \toprule
    & \multicolumn{2}{c}{Added parameters} &
      \multicolumn{3}{c}{Memory (GiB)} \\
    \cmidrule(lr){2-3}\cmidrule(l){4-6}
    Method & Count & Overhead & Params & Cache  & Cache \\
      &  &  &  &  (1K) &  (16K) \\
    \midrule
    MHA & --  & -- & $15.26$ & $0.14$ & $2.25$ \\
    Linear & $9.09$M & $0.11\%$ & $15.27$ &$0.08$ & $0.08$ \\
    \ours{} & $9.74$M & $0.12\%$ &$15.28$ & $0.12$ & $1.39$ \\
    \revised{(merged)} \\
    \revised{\ours{}} & \revised{$47.49$M} & \revised{$0.58\%$} &\revised{$15.42$} & \revised{$0.12$} & \revised{$1.39$} \\
     \revised{(unmerged)}\\
    \bottomrule
  \end{tabular}
  \vspace{-0.8em}
\end{wraptable}

Table~\ref{tab:memory-overhead} summarizes the estimated memory and parameter overhead. The linear path adds $9.09$M parameters ($0.11\%$ of the original Qwen3-8B), while the RL router adds only $0.65$M parameters ($0.0079\%$). \revised{The merged row assumes that the LoRA adapter of a layer is added to its output projection when the layer switches. This is possible because a layer never switches back to full attention (Section~\ref{sec:router}). Without merging, the LoRA adapters add $37.75$M parameters, so \ours{} adds $47.49$M parameters ($0.58\%$) in total.} Memory estimates include model weights and attention state; \ours{} uses only MATH-500 routing decisions from the accuracy-scaled break-even-cosine router. At a layer's switching step, we count both its accumulated full-attention KV cache and its linear state before releasing the KV cache. \revised{The estimates exclude activation and workspace memory and implementation buffers, such as delayed linear inputs and convolution states.} At 16K tokens, \ours{} reduces the estimated memory by $0.86$ GiB per sequence relative to full attention while adding only \revised{$0.58\%$} parameters.

\revised{On all 500 MATH-500 problems (Figure~\ref{fig:measured-cost}), \ours{} decodes longer sequences (from 8K to 16K tokens) $33\%$ faster. Its highest peak memory is $17.4$ GiB, compared with $17.9$ GiB for full attention.}

%% file: text/05_conclusion.tex
\section{Conclusion}

We introduced \ours{}, a hybrid post-hoc linearization approach that learns when each layer should switch from full to linear attention during generation. Across reasoning benchmarks, learned routing generally provides a better accuracy--memory trade-off than fixed switching, while our analysis shows that the reward design determines whether the learned policy is task-dependent or largely static.

%% file: text/appendix.tex
\newpage
\section{Linearization}
\label{app:linearization}
\subsection{Linearization Training and Objective Ablations}
\label{app:linearization-training}
\label{app:linearization-objective-ablation}

The linearized model uses the short-convolution and LoRA-on-$o_{\mathrm{proj}}$ architecture described in Section~\ref{sec:linearization}. Let $x_{1:N}$ be a training sequence, with teacher and student logits $z_t^{\mathrm{T}}$ and $z_t^{\mathrm{S}}$. We evaluate next-token prediction, logit and activation distillation loss function. 

Next-token prediction is a cross entropy loss
\begin{equation}
  \mathcal{L}_{\mathrm{CE}} = -\frac{1}{N-1}\sum_{t=1}^{N-1}
  \log p_{\mathrm{S}}(x_{t+1}\mid x_{\leq t}),
\end{equation}

Logit distillation uses KL-divergence between teacher and student token distributions. With $p_t^{\mathrm{T}}=\operatorname{softmax}(z_t^{\mathrm{T}}/\tau)$ and $p_t^{\mathrm{S}}=\operatorname{softmax}(z_t^{\mathrm{S}}/\tau)$, the loss is
\begin{equation}
  \mathcal{L}_{\mathrm{KL}} = \frac{\tau^2}{N-1}\sum_{t=1}^{N-1}
  \operatorname{KL}\!\left(p_t^{\mathrm{T}}\,\middle\|\,p_t^{\mathrm{S}}\right),
\end{equation}
where $\tau$ is a temperature hyperparameter.

For activation distillation (AD)~\citep{zhang2024lolcats}, $h_{\ell,t}^{\mathrm{T}}$ and $h_{\ell,t}^{\mathrm{S}}$ are the teacher and student activations at layer $\ell$. We use
\begin{equation}
  \mathcal{L}_{\mathrm{AD}} = \frac{1}{LN}
  \sum_{\ell\i=1}^L\sum_{t=1}^{N}
  \left\|\frac{h_{\ell,t}^{\mathrm{S}}-h_{\ell,t}^{\mathrm{T}}}
  {\sigma_{\mathrm{global}}}\right\|_2^2,
  \quad
  \sigma_{\mathrm{global}}=\operatorname{Std}\!\left(\{h_{\ell,t}^{\mathrm{T}}\}_{\ell,t}\right).
\end{equation}
The same normalizing scalar $\sigma_{\mathrm{global}}$ is shared across layers and positions. Teacher forcing uses the teacher hidden state as an input to the student layer; otherwise the student uses its own upstream activation. 

The mix of logit and activation distillation we use the weighted sum of two losses:
\begin{equation}
  \mathcal{L}_{\mathrm{mix}}=
  \lambda_{\mathrm{logit}}\mathcal{L}_{\mathrm{KL}}+
  \lambda_{\mathrm{AD}}\mathcal{L}_{\mathrm{AD}}.
\end{equation}
We use $\lambda_{\mathrm{logit}} =  \lambda_{\mathrm{AD}} = 0.5$ in our experiments.

Table~\ref{tab:linearization-objective-ablation} compares next-token prediction, logit distillation, activation distillation, teacher forcing, and LoRA adapter rank. The full model is a reference, not a linearized variant.

\begin{table}[h]
  \centering
  \scriptsize
  \setlength{\tabcolsep}{2.7pt}
  \caption{Objective and LoRA ablation using DCLM-Edu training dataset. The objective ablation uses 0.08B training tokens; the LoRA ablation uses 0.4B training tokens. $\checkmark$ marks a loss component that is used. NTP denotes next-token prediction, Logit denotes logit distillation, AD denotes activation distillation, and TF denotes teacher forcing. LAMBADA denotes LAMBADA OpenAI. LAMBADA and MMLU use no chat template or thinking; GSM8K 0-shot and MATH-500 use the chat template with thinking.}
  \label{tab:linearization-objective-ablation}
  \begin{tabular}{@{}ccccccccccc@{}}
    \toprule
    \multicolumn{4}{c}{Loss components} & \multirow{2}{*}{LoRA rank} & \multicolumn{3}{c}{Non-thinking} & \multicolumn{3}{c}{Thinking} \\
    \cmidrule(lr){1-4}\cmidrule(lr){6-8}\cmidrule(lr){9-11}
    NTP & Logit & AD & TF & & LAMBADA & MMLU & Avg. & GSM8K & MATH-500 & Avg. \\
    \midrule
    \multicolumn{5}{@{}l}{Full model} & 64.2 & 74.9 & 69.6 & 95.6 & 95.0 & 95.3 \\
    \midrule
    \multicolumn{11}{@{}l}{\textit{Objective ablation}} \\
    \checkmark & & & & 128 & \textbf{67.0} & 70.4 & \textbf{68.7} & 8.0 & 8.0 & 8.0 \\
    & \checkmark & & & 128 & 62.9 & 72.3 & 67.6 & 14.7 & \textbf{15.2} & \textbf{15.0} \\
    & & \checkmark & & 128 & 61.6 & 72.4 & 67.0 & 5.0 & 5.8 & 5.4 \\
    & & \checkmark & \checkmark & 128 & 62.1 & 70.7 & 66.4 & 2.3 & 1.8 & 2.1 \\
    & \checkmark & \checkmark & & 128 & 62.4 & \textbf{72.6} & 67.5 & \textbf{15.6} & 13.8 & 14.7 \\
    \midrule
    \multicolumn{11}{@{}l}{\textit{Adapter ablation (mixed Logit+AD)}} \\
    & \checkmark & \checkmark & & -- & 62.5 & \textbf{73.2} & \textbf{67.9} & 8.3 & 10.6 & 9.5 \\
    & \checkmark & \checkmark & & 128 & \textbf{62.7} & 72.8 & 67.8 & \textbf{30.4} & \textbf{20.8} & \textbf{25.6} \\
    & \checkmark & \checkmark & & 256 & 62.6 & 73.0 & 67.8 & 28.1 & \textbf{20.8} & 24.5 \\
    \bottomrule
  \end{tabular}
\end{table}

\newpage
\subsection{Training-Data Ablation}
\label{app:linearization-data-ablation}

We compare three training mixtures: DCLM-Edu (DCLM), OpenThoughts3 (OT), and DCLM followed by OT. The combined starts with 1B DCLM-Edu tokens and then continues on OT until reaching the reported total token count. We report accuracies across all the benchmarks in Table~\ref{tab:linearization-data-ablation}. All the experiments use a mix of logit and activation distillation objective without teacher forcing. Sliding window size if fixed at 64, 8 sink tokens and LoRA rank 128.

\begin{table*}[h]
  \centering
  \small
  \setlength{\tabcolsep}{3.2pt}
  \caption{Training-data ablation for the linearized model. LAMBADA and MMLU use no chat template or thinking; GSM8K 0-shot and MATH-500 use long-thinking evaluation. GPQA denotes GPQA-Diamond, and HE+ denotes HumanEval+. Each Avg. column is the unweighted mean within its benchmark group. For Llama~3.1-8B, we use one linearized checkpoint trained on 4.5B OpenThoughts tokens.}
  \label{tab:linearization-data-ablation}
  \resizebox{\textwidth}{!}{%
  \begin{tabular}{@{}lcccccccccccc@{}}
    \toprule
    & & & & \multicolumn{3}{c}{Non-thinking} & \multicolumn{6}{c}{Thinking} \\
    \cmidrule(lr){5-7}\cmidrule(lr){8-13}
    Model (training data) & SWA & Sink & Tok. & LAMBADA & MMLU & Avg. & GSM8K & MATH-500 & GPQA & HE+ & MBPP+ & Avg. \\
    \midrule
    \multicolumn{13}{@{}l}{\textit{Qwen3-8B}} \\
    Full model & -- & -- & -- & 64.2 & 74.9 & 69.6 & 95.6 & 95.0 & 56.6 & 84.8 & 93.4 & 85.1 \\
    SWA-only & 64 & 8 & -- & 62.0 & 70.9 & 66.5 & 1.8 & 3.0 & 3.5 & 0.0 & 0.0 & 1.7 \\
    \cmidrule(lr){1-13}
    \ours{} (DCLM) & 64 & 8 & 0.4B & 62.7 & 72.8 & 67.8 & 30.4 & 20.8 & 16.2 & 17.1 & 22.5 & 21.4 \\
    \ours{} (DCLM) & 64 & 8 & 1.0B & \textbf{63.0} & 73.0 & 68.0 & 33.7 & 25.2 & 16.7 & 22.0 & 23.5 & 24.2 \\
    \ours{} (DCLM) & 64 & 8 & 2.0B & \textbf{63.0} & \textbf{73.3} & \textbf{68.2} & 36.1 & 26.2 & 18.7 & 16.5 & 23.8 & 24.3 \\
    \cmidrule(lr){1-13}
    \ours{} (OT) & 64 & 8 & 0.5B & 61.1 & 70.7 & 65.9 & 64.6 & 53.4 & 23.2 & 36.0 & 27.0 & 40.8 \\
    \ours{} (OT) & 64 & 8 & 1.0B & 62.4 & 70.0 & 66.2 & 69.4 & 52.8 & \textbf{28.3} & \textbf{36.6} & 29.4 & 43.3 \\
    \ours{} (OT) & 64 & 8 & 2.0B & 60.8 & 69.2 & 65.0 & 73.8 & 60.6 & 23.2 & 34.1 & 27.2 & 43.8 \\
    \ours{} (OT) & 64 & 8 & 4.5B & 57.8 & 67.9 & 62.9 & \textbf{79.5} & \textbf{64.0} & 27.3 & 34.8 & 24.1 & \textbf{45.9} \\
    \cmidrule(lr){1-13}
    \ours{} (DCLM $\rightarrow$ OT) & 64 & 8 & 2.0B & 62.6 & 69.9 & 66.3 & 71.7 & 56.2 & 20.2 & 29.9 & \textbf{33.1} & 42.2 \\
    \ours{} (DCLM $\rightarrow$ OT) & 64 & 8 & 3.0B & 62.1 & 69.1 & 65.6 & 74.5 & 61.2 & 19.2 & 29.9 & 27.5 & 42.5 \\
    \specialrule{1pt}{2pt}{2pt}
    \multicolumn{13}{@{}l}{\textit{Llama 3.1-8B}} \\
    Full model & -- & -- & -- & 72.2 & 68.3 & 70.3 & 78.2 & 48.6 & 23.7 & 58.5 & 73.0 & 56.4 \\
    SWA-only & 64 & 8 & -- & 69.4 & 64.3 & 66.9 & 3.0 & 6.8 & 4.0 & 11.0 & 38.1 & 12.6 \\
    \ours{} (OT) & 64 & 8 & 4.5B & 67.0 & 58.8 & 62.9 & 51.2 & 28.0 & 8.1 & 25.6 & 31.2 & 28.8 \\
    \bottomrule
  \end{tabular}%
  }
\end{table*}

\clearpage
\newpage
\subsection{Sliding-Window Size Ablation}
\label{app:swa-window-ablation}

Figure~\ref{fig:swa-window-ablation} presents the Qwen3-8B and Llama~3.1-8B sliding-window comparisons. We report the values used to produce these figure in Tables~\ref{tab:swa-window-ablation} and~\ref{tab:llama-swa-window-ablation}. All variants use the evaluation protocols defined in Section~\ref{sec:linearization}. For each model, the trained SWA-64 result is the 4.5B-token OpenThoughts checkpoint from the training-data ablation. SWA-128 and SWA-512 evaluate the same checkpoint with a different SWA size.

\begin{table*}[h]
  \centering
  \small
  \setlength{\tabcolsep}{3.0pt}
  \caption{Qwen3-8B sliding-window ablation corresponding to Figure~\ref{fig:swa-window-ablation}(a). LAMBADA and MMLU use no chat template or thinking; GSM8K 0-shot and MATH-500 use long-thinking evaluation. GPQA uses the Qwen-thinking task and final-answer metric. HE+ denotes HumanEval+; HE+ and MBPP+ report pass@1. Each Avg. column is the unweighted mean within its benchmark group.}
  \label{tab:swa-window-ablation}
  \resizebox{\textwidth}{!}{%
  \begin{tabular}{@{}lcccccccccccc@{}}
    \toprule
    & & & & \multicolumn{3}{c}{Non-thinking} & \multicolumn{6}{c}{Thinking} \\
    \cmidrule(lr){5-7}\cmidrule(lr){8-13}
    Model (training data) & SWA & Sink & Tok. & LAMBADA & MMLU & Avg. & GSM8K & MATH-500 & GPQA & HE+ & MBPP+ & Avg. \\
    \midrule
    Full model & -- & -- & -- & 64.2 & 74.9 & 69.6 & 95.6 & 95.0 & 56.6 & 84.8 & 93.4 & 85.1 \\
    \cmidrule(lr){1-13}
    SWA-only & 64 & 8 & -- & 62.0 & 70.9 & 66.5 & 1.8 & 3.0 & 3.5 & 0.0 & 0.0 & 1.7 \\
    SWA-only & 128 & 8 & -- & 64.1 & 74.1 & 69.1 & 9.2 & 9.0 & 3.0 & 0.0 & 0.3 & 4.3 \\
    SWA-only & 512 & 8 & -- & 64.1 & 74.9 & 69.5 & 74.1 & 47.4 & 12.6 & 6.7 & 8.7 & 29.9 \\
    \cmidrule(lr){1-13}
    \ours{} (OT) & 64 & 8 & 4.5B & 57.8 & 67.9 & 62.9 & 79.5 & 64.0 & 27.3 & 34.8 & 24.1 & 45.9 \\
    \ours{} (OT) & 128 & 8 & 4.5B & 59.1 & 69.6 & 64.4 & 83.9 & 68.4 & 27.8 & 37.8 & 28.0 & 49.2 \\
    \ours{} (OT) & 512 & 8 & 4.5B & 59.1 & 69.8 & 64.5 & 90.7 & 76.8 & 28.3 & 48.8 & 50.8 & 59.1 \\
    \bottomrule
  \end{tabular}%
  }
\end{table*}

\begin{table*}[h]
  \centering
  \small
  \setlength{\tabcolsep}{3.0pt}
  \caption{Llama~3.1-8B sliding-window ablation corresponding to Figure~\ref{fig:swa-window-ablation}(b). LAMBADA and MMLU use no chat template or thinking; GSM8K 0-shot and MATH-500 use long-thinking evaluation. GPQA denotes GPQA-Diamond, and HE+ denotes HumanEval+. HE+ and MBPP+ report pass@1. Each Avg. column is the unweighted mean within its benchmark group.}
  \label{tab:llama-swa-window-ablation}
  \resizebox{\textwidth}{!}{%
  \begin{tabular}{@{}lcccccccccccc@{}}
    \toprule
    & & & & \multicolumn{3}{c}{Non-thinking} & \multicolumn{6}{c}{Thinking} \\
    \cmidrule(lr){5-7}\cmidrule(lr){8-13}
    Model (training data) & SWA & Sink & Tok. & LAMBADA & MMLU & Avg. & GSM8K & MATH-500 & GPQA & HE+ & MBPP+ & Avg. \\
    \midrule
    Full model & -- & -- & -- & 72.2 & 68.3 & 70.3 & 78.2 & 48.6 & 23.7 & 58.5 & 73.0 & 56.4 \\
    \cmidrule(lr){1-13}
    SWA-only & 64 & 8 & -- & 69.4 & 64.3 & 66.9 & 3.0 & 6.8 & 4.0 & 11.0 & 38.1 & 12.6 \\
    SWA-only & 128 & 8 & -- & 72.2 & 66.5 & 69.4 & 27.7 & 18.4 & 3.0 & 31.1 & 65.6 & 29.2 \\
    SWA-only & 512 & 8 & -- & 72.2 & 68.1 & 70.2 & 78.5 & 49.0 & 16.7 & 57.9 & 74.9 & 55.4 \\
    \cmidrule(lr){1-13}
    \ours{} (OT) & 64 & 8 & 4.5B & 67.0 & 58.8 & 62.9 & 51.2 & 28.0 & 8.1 & 25.6 & 31.2 & 28.8 \\
    \ours{} (OT) & 128 & 8 & 4.5B & 68.8 & 60.1 & 64.5 & 65.0 & 36.8 & 9.6 & 36.0 & 45.2 & 38.5 \\
    \ours{} (OT) & 512 & 8 & 4.5B & 68.8 & 61.3 & 65.1 & 72.4 & 44.2 & 16.7 & 45.7 & 62.2 & 48.2 \\
    \bottomrule
  \end{tabular}%
  }
\end{table*}

\clearpage
\subsection{Linearization Hyperparameters}
\label{app:linearization-hyperparameters}

\begin{table*}[h]
  \centering
  \small
  \setlength{\tabcolsep}{6pt}
  \renewcommand{\arraystretch}{1.02}
  \caption{Linearization training and evaluation hyperparameters used for both Qwen3-8B and Llama~3.1-8B.}
  \label{tab:linearization-hyperparameters}
  \begin{tabular}{@{}p{0.43\textwidth}p{0.51\textwidth}@{}}
    \toprule
    \textbf{Hyperparameter} & \textbf{Value} \\
    \midrule
    \multicolumn{2}{c}{\textbf{Training data}} \\
    Teacher / backbone $f_{\mathrm{T}}$ & \texttt{Qwen/Qwen3-8B}; \texttt{Meta-Llama-3.1-8B-Instruct} \\
    Dataset $\mathcal{D}$ & OpenThoughts3-1.2M (train) \\
    Input formatting & Each model's chat template \\
    Tokens used in the loss & Assistant-response tokens only \\
    Gradient propagation length $C$ & $4096$ tokens \\
    Input-token budget & $4.5$B ($0.25$ epoch) \\
    \midrule
    \multicolumn{2}{c}{\textbf{Linear path}} \\
    Linear mixer & GDN in every attention layer \\
    SWA window $W$ & $64$ \\
    Sink tokens $s$ & $8$ \\
    Gate dimensions & Scalar key gate; scalar $\beta$ \\
    $\beta$ activation & Softplus \\
    Gate initialization & $b_g=1$, $b_\beta=-1.5$ \\
    Short convolution & $k=8$, bias $=\mathrm{True}$ \\
    Trainable parameters & GDN, short convolution, and LoRA; backbone frozen \\
    \midrule
    \multicolumn{2}{c}{\textbf{LoRA}} \\
    Target module & $o_{\mathrm{proj}}$ \\
    Rank $r$ & $128$ \\
    Scale $\alpha$ & $8$ \\
    Dropout $p$ & $0$ \\
    \midrule
    \multicolumn{2}{c}{\textbf{Objective}} \\
    Loss weights $(\lambda_{\mathrm{KL}},\lambda_{\mathrm{AD}})$ & $(0.5,0.5)$ \\
    KL direction & $\operatorname{KL}(p_{\mathrm{T}}\,\|\,p_{\mathrm{S}})$ \\
    Distillation temperature $\tau$ & $2$ \\
    Activation loss & Squared $\ell_2$ \\
    Teacher forcing & False \\
    \midrule
    \multicolumn{2}{c}{\textbf{Optimization}} \\
    Optimizer & AdamW \\
    Learning rate $\eta$ & $10^{-3}$ \\
    Adam $(\beta_1,\beta_2)$ & $(0.9,0.95)$ \\
    Adam $\epsilon$ & $10^{-8}$ \\
    Weight decay & $0$ \\
    LR schedule & Cosine \\
    Warmup steps & $100$ \\
    Gradient clipping & $1$ \\
    Training steps $S$ & $150{,}000$ \\
    Batch size per GPU & $1$ \\
    Number of GPUs & $2$ \\
    Gradient accumulation & $1$ \\
    Precision & bfloat16 \\
    Seed & $1$ \\
    \midrule
    \multicolumn{2}{c}{\textbf{Generation (reasoning and code)}} \\
    Sampling & True \\
    Temperature $T_{\mathrm{gen}}$ & $0.6$ \\
    Top-$p$ & $0.95$ \\
    Top-$k$ & $20$ \\
    Maximum generated tokens $G_{\max}$ & $16{,}384$ \\
    Samples per prompt & $1$ \\
    \bottomrule
  \end{tabular}
\end{table*}

%%%%%%%%%%%%%%%%%%%%%%%%%%%%%%%%%%%%%%%%%%%%%%%%%%%%%%%%%%%%%%%
% RL related appendix
%%%%%%%%%%%%%%%%%%%%%%%%%%%%%%%%%%%%%%%%%%%%%%%%%%%%%%%%%%%%%%%
\clearpage
\section{Reinforcement Learning}
\label{app:reinforcement-learning}

\subsection{Router Architecture}
\label{app:router-architecture}

\paragraph{Schedule}
Let $\boldsymbol{\psi}(t)$ be the log-spaced sinusoidal encoding of
position $t$, with $F$ frequencies and maximum period $\Omega$:
\begin{equation}
  \boldsymbol{\psi}(t)=
  \left[
    \sin\left(t\Omega^{-i/(F-1)}\right);
    \cos\left(t\Omega^{-i/(F-1)}\right)
  \right]_{i=0}^{F-1}, \qquad \Omega=10{,}000.
\end{equation}

The positional input in Section~\ref{sec:router} is
\begin{equation}
  \mathbf x^{\mathrm{schedule}}_t=
  \left[
    \boldsymbol{\psi}(t);
    \frac{t}{B};
    \operatorname{clip}\left(\frac{t}{B},0,1\right);
    \max\left(\frac{t-B}{B},0\right);
    \mathbf m_{t-1}
  \right],
  \qquad \mathbf m_{t-1}\in\{0,1\}^{L}.
  \label{eq:router-architecture-input}
\end{equation}

This vector has $2F+3+L$ entries. We use a per-layer MLP to produce the schedule-dependent part of the router:
\begin{equation}
\begin{aligned}
  \nu^{\mathrm{schedule}}_{\ell,t}&=
    W^{\mathrm{out}}_\ell\operatorname{SiLU}\!\left(
      W^{\mathrm{in}}_\ell\mathbf x^{\mathrm{schedule}}_t+
      \mathbf b^{\mathrm{in}}_\ell\right)+b^{\mathrm{out}}_\ell.\\
  W^{\mathrm{in}}_\ell& \in \mathbb{R}^{H^{\mathrm{schedule}}\times(2F+3+L)},
  \quad \mathbf b^{\mathrm{in}}_\ell\in\mathbb{R}^{H^{\mathrm{schedule}}},
  \quad W^{\mathrm{out}}_\ell\in\mathbb{R}^{1\times H^{\mathrm{schedule}}},
  \quad b^{\mathrm{out}}_\ell\in\mathbb{R}.
  \end{aligned}
\end{equation}    \label{eq:router-architecture-schedule}
We initialize $W^{\mathrm{in}}_\ell$ with Xavier uniform, set $\mathbf b^{\mathrm{in}}_\ell=0$ and $W^{\mathrm{out}}_\ell=0$, and set $b^{\mathrm{out}}_\ell = 6$. \revised{The router computes the stay-full probability in bfloat16, which rounds it to $1-1/256\approx 99.6\%$. Therefore, at initialization, each layer switches, on average, at token 256.}

\paragraph{Context}
The context input is the hidden representation of the corresponding attention layer:
\[
  \operatorname{sg}[\mathbf h_{\ell,t}]\in\mathbb{R}^{H^{\mathrm{model}}}.
\]
Here $H^{\mathrm{model}}$ is the hidden size of the frozen language model. Because it is typically large, the first layer of the context MLP is shared across the LLM layers to save router parameters. Only its readout is layer-specific, so it carries the index $\ell$:
\begin{equation}
\begin{aligned}
  \nu^{\mathrm{context}}_{\ell,t}&=
    W_\ell\operatorname{SiLU}\left(
      W^{\mathrm{c}}\operatorname{sg}[\mathbf h_{\ell,t}]+\mathbf b^{\mathrm{c}}\right).\\
  W^{\mathrm{c}}&\in\mathbb{R}^{H^{\mathrm{context}}\times H^{\mathrm{model}}},
  \quad \mathbf b^{\mathrm{c}}\in\mathbb{R}^{H^{\mathrm{context}}},
  \quad W_\ell\in\mathbb{R}^{1\times H^{\mathrm{context}}}.
  \label{eq:router-architecture-context}
\end{aligned}
\end{equation}
We initialize $W^{\mathrm{c}}$ with Xavier uniform, set $\mathbf b^{\mathrm{c}}=0$, and set $W_\ell=0$.

The total score is $\nu_{\ell,t}=\nu^{\mathrm{schedule}}_{\ell,t}+
\nu^{\mathrm{context}}_{\ell,t}$.

Table~\ref{tab:router-training-hyperparameters} specifies the corresponding architecture hyperparameters.

\newpage
\subsection{\revised{Switching Implementation}}
\label{app:switching-implementation}
\revised{We describe one attention head of layer $\ell$ with switch position $\tau=\tau_\ell$. Let $h_t$ be the input of the layer at token $t$, and let $q_t$, $k_t$, and $v_t$ be the query, key, and value of full attention. The KV cache stores $k_t$ and $v_t$, and the scalar gates $\gamma_t\in(0,1)$ and $\beta_t>0$ are computed from $h_t$. The first $n_s=8$ tokens are sink tokens and do not enter the GDN state. The GDN state after tokens $n_s<j\leq n$ depends only on the cached keys and values and on the gates:
\begin{equation}
  S_n=\mathrm{GDN}\bigl(k_{1:n},\,v_{1:n},\,\gamma_{n_s+1:n},\,\beta_{n_s+1:n}\bigr).
  \label{eq:gdn-state}
\end{equation}
For the gated delta rule~\citep{yang2024gated} with short convolutions of kernel size $c=8$, the state is computed as follows:
\begin{equation}
\begin{aligned}
  \tilde k_j&=\mathrm{Conv}_k(k_{j-c+1:j}),\qquad
  \hat k_j=\tilde k_j/\lVert\tilde k_j\rVert_2,\qquad
  \tilde v_j=\mathrm{Conv}_v(v_{j-c+1:j}),\\
  \Gamma_j&=\prod_{i=n_s+1}^{j}\gamma_i,\\
  u_j&=\beta_j\Bigl(\tilde v_j-\sum_{i=n_s+1}^{j-1}\frac{\Gamma_j}{\Gamma_i}\bigl(\hat k_j^{\top}\hat k_i\bigr)u_i\Bigr),\qquad n_s<j\leq n,\\
  S_n&=\sum_{j=n_s+1}^{n}\frac{\Gamma_n}{\Gamma_j}\,u_j\hat k_j^{\top}.
\end{aligned}
\label{eq:gdn-compute}
\end{equation}
The vectors $u_j$ are the solution of a lower-triangular linear system, so $S_n$ is computed from all tokens at once. A token enters the state only after it leaves the sliding window of $w=64$ tokens. The output of the linear path at token $t$ is
\begin{equation}
  o_t=S_{t-w}\,\hat q_t+\mathrm{Attn}\bigl(q_t;\,\{1,\ldots,n_s\}\cup\{t-w+1,\ldots,t\}\bigr),
\end{equation}
where $\hat q_t$ is the convolved and normalized query and $\mathrm{Attn}(q_t;\mathcal{I})$ is softmax attention over the cached tokens in the set $\mathcal{I}$.}

\revised{\textbf{Switching.} For $t\leq\tau$, the layer uses full attention. Besides $k_t$ and $v_t$ in the KV cache, we store $\gamma_t$ and $\beta_t$, which are two scalars per head and token. At the switch, we compute $S_{\tau-w}$ with Equations~\ref{eq:gdn-state} and~\ref{eq:gdn-compute} from the cached $(k_j, v_j, \gamma_j, \beta_j)$ for $n_s<j\leq\tau-w$. We solve the linear system for $u_j$ with the chunked GDN kernel of the Flash Linear Attention library~\citep{yang2024fla}, with its default chunk size of 64 tokens. The convolution history is the last $c-1$ cached keys and values. All these quantities are the same as the linear path would compute from the inputs $h_{1:\tau}$. Therefore, $S_{\tau-w}$ is exactly the state of the linear path after the same prefix, and the switch does not need extra training or approximation. For $t>\tau$, the layer uses the linear path and adds each token to the state with the standard GDN decoding update when the token leaves the sliding window. We release the KV cache of tokens $n_s<j\leq\tau-w$ and keep only the $n_s$ sink tokens and the last $w$ tokens.}

\newpage
\subsection{Accuracy--Memory Results}
\label{app:accuracy-memory-results}

Table~\ref{tab:accuracy-memory-results} reports the numerical values used to produce Figure~\ref{fig:hybridization-memory-accuracy}.

\begin{table*}[h]
  \centering
  \caption{Accuracy and relative cache saving for the configurations in Figure~\ref{fig:hybridization-memory-accuracy}.}
  % Data source: poirot/notebooks/iclr_results/04_routing_decisions.ipynb, cells 7 and 8.
  \label{tab:accuracy-memory-results}
  \resizebox{\textwidth}{!}{%
  \begin{tabular}{@{}cccccccccccc@{}}
    \toprule
    \multicolumn{5}{c}{Router configuration} & \multicolumn{6}{c}{Accuracy} & Cache saving \\
    \cmidrule(lr){1-5}\cmidrule(lr){6-11}\cmidrule(l){12-12}
    Reward shape & Efficiency score & Context & $\Delta$ & $\alpha$ & GSM8K & GPQA-D & MATH-500 & MBPP+ & HE+ & Avg. & (\%) $\uparrow$ \\
    \midrule
    \multicolumn{5}{l}{Full attention} & 95.6 & 56.6 & 95.0 & 93.4 & 84.8 & 85.1 & 0.0 \\
    \multicolumn{5}{l}{All linear} & 79.5 & 27.3 & 64.0 & 24.1 & 34.8 & 45.9 & 69.4 \\
    \midrule
    \multicolumn{5}{l}{Per-layer hybridization (10\% full layers)} & 82.4 & 25.8 & 67.6 & 20.9 & 40.9 & 47.5 & 61.7 \\
    \multicolumn{5}{l}{Per-layer hybridization (25\% full layers)} & 87.0 & 24.7 & 75.4 & 22.0 & 37.8 & 49.4 & 52.1 \\
    \multicolumn{5}{l}{Per-layer hybridization (50\% full layers)} & 93.2 & 39.4 & 89.4 & 73.5 & 77.4 & 74.6 & 34.7 \\
    \multicolumn{5}{l}{Manual switching (1K tokens)} & 79.8 & 30.3 & 58.0 & 26.5 & 21.3 & 43.2 & 50.8 \\
    \multicolumn{5}{l}{Manual switching (2K tokens)} & 91.4 & 33.8 & 64.6 & 56.9 & 50.6 & 59.5 & 27.9 \\
    \multicolumn{5}{l}{Manual switching (4K tokens)} & 94.7 & 39.9 & 75.6 & 81.2 & 75.6 & 73.4 & 12.0 \\
    \midrule
    Acc.-scaled & Break-even cos. & $\checkmark$ & 8K & 0.1 & 95.3 & 50.0 & 93.0 & 91.3 & 82.3 & 82.4 & 25.4 \\
    Additive & Break-even cos.& $\checkmark$ & 8K & 0.1 & 95.9 & 55.1 & 94.8 & 93.4 & 76.2 & 83.1 & 14.4 \\
    Acc.-scaled & Memory-only & $\checkmark$ & 8K & 0.1 & 94.2 & 47.0 & 88.4 & 79.9 & 63.4 & 74.6 & 39.3 \\
    \bottomrule
  \end{tabular}%
  }
\end{table*}

\subsection{\revised{FlashMorph Baseline}}
\label{app:flashmorph}
\revised{FlashMorph~\citep{lan2026morphing} builds a per-layer hybrid model from a full-attention model and its linear counterpart. Each layer $\ell$ has a scalar gate $a_\ell\in[0,1]$, and the attention output of the layer is $a_\ell\,o^{\mathrm{full}}_\ell+(1-a_\ell)\,o^{\mathrm{lin}}_\ell$. All model weights stay frozen and only the gates are trained. The loss is the mean squared error between the hidden states of the gated model and of the full-attention model, averaged over all layers and computed on response tokens, plus the sparsity penalty $\lambda\sum_\ell a_\ell$. After training, the layers with the largest gates keep full attention and all other layers use the linear path.}

\revised{No official code is available, so we reimplemented the method. We use the full-attention and linear paths of Section~\ref{sec:hybrid-experiments} and follow the settings of the original paper: all gates start at $1$, $\lambda=0.1$, $250$ steps with $8$ sequences per step, and AdamW with a warmup-stable-decay schedule from $2\times10^{-2}$ to $2\times10^{-3}$. We clip the gates to $[0,1]$ after each step. The training sequences are $2{,}000$ router-training prompts with their correct Qwen3-8B responses, truncated to $8{,}192$ tokens. The original method also fine-tunes the hybrid model after selecting the layers. We skip this step, so FlashMorph and our per-layer hybridization differ only in which layers keep full attention. We evaluate both in the same way.}

\revised{\citet{lan2026morphing} also report the layer ranking that FlashMorph learns for Qwen3-8B with their own linear layers. We evaluate this published ranking as a second baseline, FlashMorph (paper), with our linear path and without fine-tuning. We call the ranking from our reimplementation FlashMorph (ours).}

\revised{Table~\ref{tab:flashmorph-layers} lists the selected layers. FlashMorph (ours) keeps the first layers in full attention, whereas our per-layer selection keeps the last layers. The two selections share no layer at $10\%$ and $25\%$ and only layer 35 at $50\%$. The published ranking spreads the full layers over the whole model. It shares 2 layers with our per-layer selection at $25\%$ and 7 layers at $50\%$.}

\begin{table}[h]
  \centering
  \caption{\revised{Layers that keep full attention in the per-layer hybrid models.}}
  \label{tab:flashmorph-layers}
  \small
  \revisioncolor
  \begin{tabular}{@{}llll@{}}
    \toprule
    Full layers & Per-layer hybridization & FlashMorph (ours) & FlashMorph (paper) \\
    \midrule
    $10\%$ (4) & 32--35 & 2, 3, 5, 7 & -- \\
    $25\%$ (9) & 24, 28--35 & 0--3, 5--7, 11, 12 & 0, 1, 3, 7, 13, 15, 22, 24, 29 \\
    $50\%$ (18) & 18--35 & 0--16, 35 & 0--3, 7--9, 12, 13, 15, 17, 19, 20, 22, 24, 29, 33, 34 \\
    \bottomrule
  \end{tabular}
\end{table}

\newpage
\subsection{Router Training Hyperparameters}
\label{app:router-training-hyperparameters}

\begin{table*}[h]
  \centering
  \small
  \setlength{\tabcolsep}{6pt}
  \renewcommand{\arraystretch}{1.02}
  \caption{Router-training hyperparameters for Table~\ref{tab:reward-ablation}.}
  \label{tab:router-training-hyperparameters}
  \begin{tabular}{@{}p{0.43\textwidth}p{0.51\textwidth}@{}}
    \toprule
    \textbf{Hyperparameter} & \textbf{Value} \\
    \midrule
    \multicolumn{2}{c}{\textbf{Router}} \\
    Schedule hidden size $H^{\mathrm{schedule}}$ & $32$ \\
    Schedule activation & SiLU \\
    Sinusoidal frequencies $F$ & $32$ \\
    Sinusoidal maximum period $\Omega$ & $10{,}000$ \\
    Context hidden size $H^{\mathrm{context}}$ & $128$ \\
    Context activation & SiLU \\
    \midrule
    \multicolumn{2}{c}{\textbf{Reward}} \\
    Efficiency weight $\alpha$ & $0.1$ \\
    Break-even point $B$ & $512$ tokens \\
    Decay $\Delta$ & $8192$ tokens \\
    \midrule
    \multicolumn{2}{c}{\textbf{Optimization}} \\
    Optimizer & AdamW \\
    Learning rate $\eta$ & $10^{-2}$ \\
    Adam ($\beta_1$, $\beta_2$) & ($0.9$, $0.999$) \\
    Adam $\epsilon$ & $10^{-8}$ \\
    Weight decay & $0$ \\
    LR schedule & Constant \\
    Warmup steps & $0$ \\
    Gradient clipping & $1$ \\
    Training steps & $100$ \\
    Group size $K$ & $4$ \\
    Batch size per GPU & $8$ \\
    Number of GPUs & $4$ \\
    Gradient accumulation & $1$ \\
    Advantage constant $\delta$ & $10^{-6}$ \\
    Precision & bfloat16 \\
    Seed & $0$ \\
    \midrule
    \multicolumn{2}{c}{\textbf{Generation}} \\
    Sampling & True \\
    Temperature & $1.0$ \\
    Top-$p$ & -- \\
    Top-$k$ & -- \\
    Maximum generated tokens & $16{,}384$ \\
    \bottomrule
  \end{tabular}
\end{table*}

\clearpage
\subsection{\revised{Response Lengths}}
\label{app:routed-lengths}
\begin{figure}[h]
  \centering
  \includegraphics[width=\textwidth]{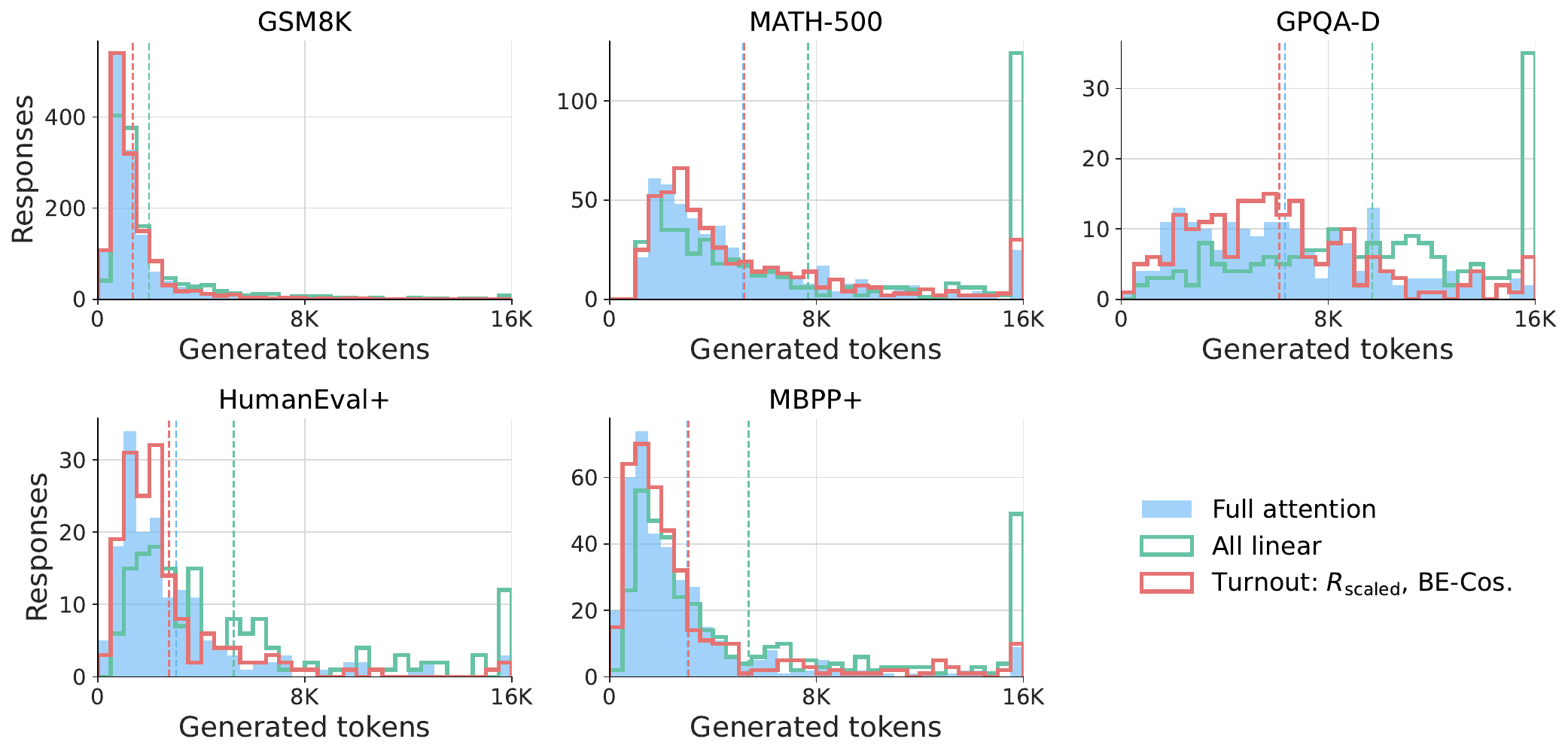}
  \vskip -5pt
  {\revisioncolor
  \caption{Histograms of generated-response lengths of full attention, the all-linear model, and \ours{} ($R_{\mathrm{scaled}}$, BE-Cos.). Dashed lines show means.}
  \label{fig:length-histograms}
  }
\end{figure}

\revised{Figure~\ref{fig:length-histograms} shows the generated-response lengths of full attention, the all-linear model, and \ours{} on the five benchmarks of Figure~\ref{fig:hybridization-memory-accuracy}. All models generate with natural stopping and a 16,384-token cap. On average, the all-linear model generates 1.5 to 1.8 times more tokens than full attention, while \ours{} and full attention generate responses of similar length.}

\clearpage
\subsection{\revised{Switching Behavior During Training}}
\label{app:router-training-dynamics}
\label{app:switch-positions}
\begin{figure}[h]
\centering
\includegraphics[width=0.75\textwidth]{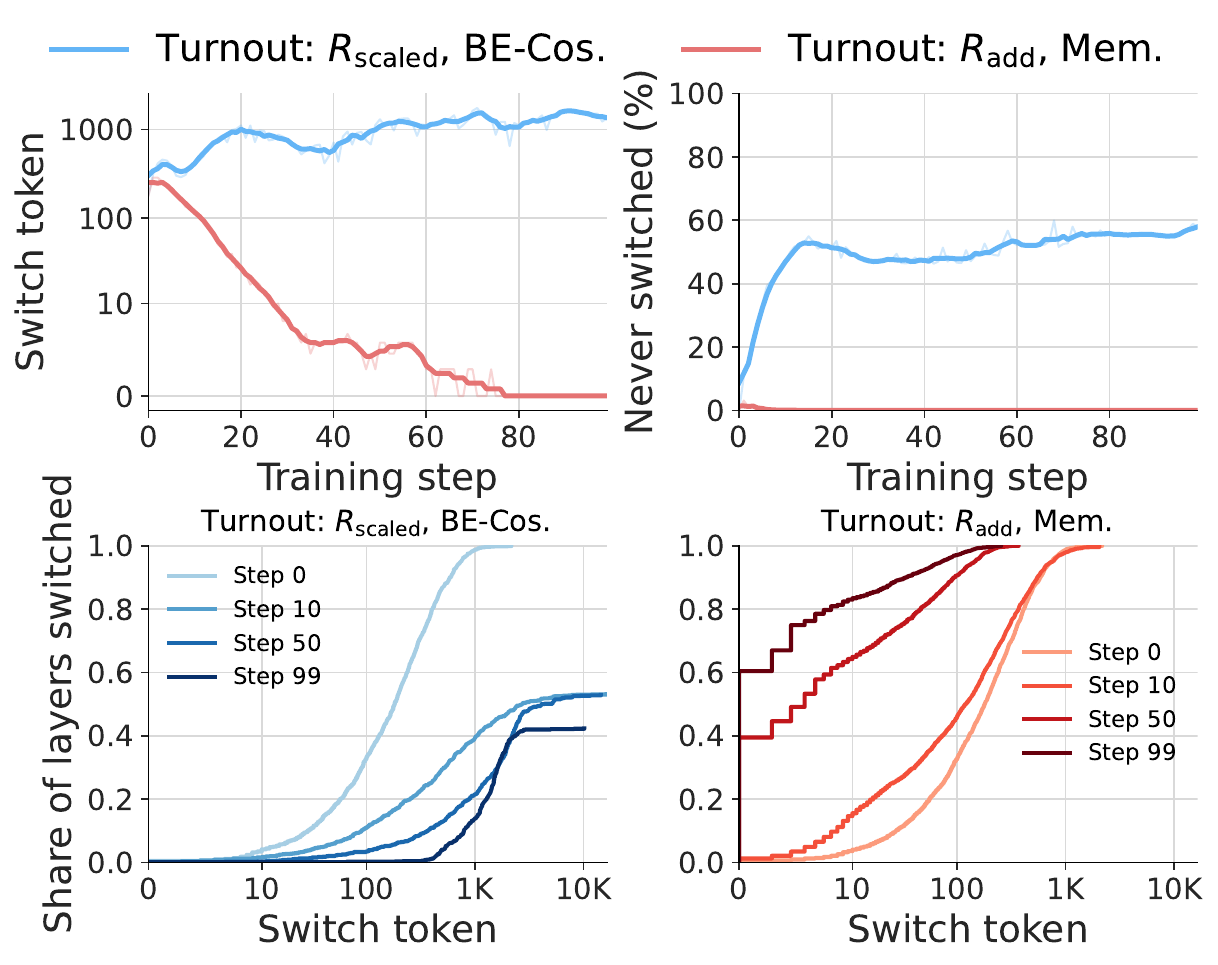}
{\revisioncolor
\caption{Training behavior of \ours{} ($R_{\mathrm{scaled}}$, BE-Cos.) and \ours{} ($R_{\mathrm{add}}$, Mem.). Top: switch token is the median number of generated tokens before a layer switches, over layers that switch before the response ends. Never switched is the share of layer--response pairs that stay in full attention. Thin lines show single steps, and thick lines show a 5-step moving average. Bottom: cumulative distribution of switch tokens over all layer--response pairs at training steps 0, 10, 50, and 99. A curve that ends below 1 means that the remaining layers never switch.}
\label{fig:router-training-dynamics}
}
\end{figure}

\revised{Figure~\ref{fig:router-training-dynamics} shows the switching positions in the training rollouts of \ours{} ($R_{\mathrm{scaled}}$, BE-Cos.) and \ours{} ($R_{\mathrm{add}}$, Mem.). At initialization, the two routers behave the same. With \ours{} ($R_{\mathrm{scaled}}$, BE-Cos.), about half of the layers stop switching within the first 10 steps, and the other layers switch later and later. With \ours{} ($R_{\mathrm{add}}$, Mem.), the router learns to switch as early as possible.}

\begin{figure}[h]
\centering
% Plot source: rl-switch-policy-rebuttal/router_layer_profile_tokens.py.
\includegraphics[width=\textwidth]{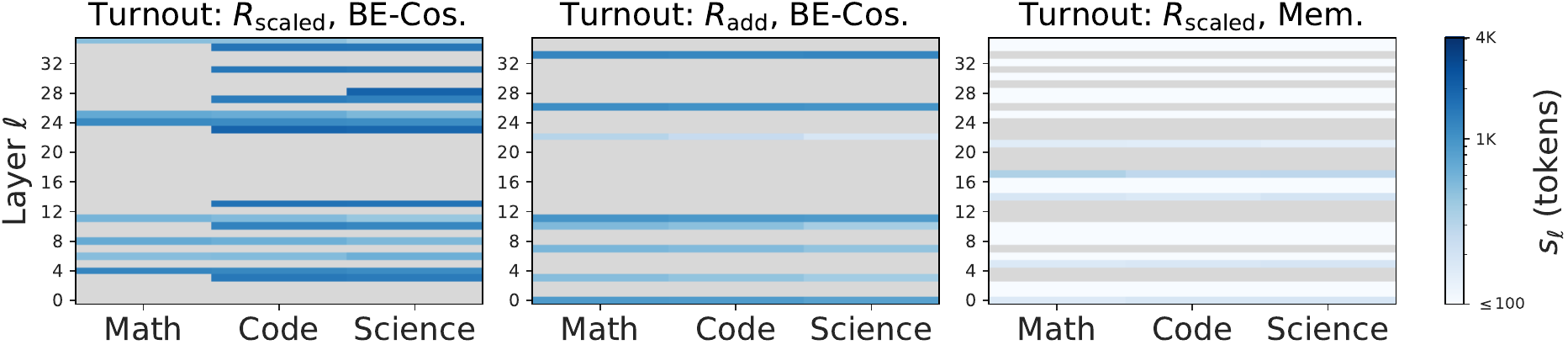}
{\revisioncolor
\caption{Absolute switching positions for the routers of Figure~\ref{fig:router-layer-profile}. Each cell shows the median switch token $s_\ell$ in generated tokens, over responses in which the layer switches. Grey cells mark layers that switch in fewer than half of the responses. Math combines GSM8K and MATH-500, Code combines HumanEval+ and MBPP+, and Science is GPQA-D.}
\label{fig:switch-positions}
}
\end{figure}

\revised{Figure~\ref{fig:switch-positions} shows the switching positions of Figure~\ref{fig:router-layer-profile} in absolute generated tokens.}

